\documentclass[10pt,twocolumn,letterpaper]{article}
\usepackage{arxiv_style}
\usepackage[T1]{fontenc}
\usepackage{tikz}
\definecolor{PlotDirect}{HTML}{A4ADB7}
\definecolor{PlotN1}{HTML}{0072B2}
\definecolor{PlotUnknown}{HTML}{DCE1E6}
\definecolor{DsAlgo}{HTML}{0072B2}
\definecolor{DsVisual}{HTML}{D48900}
\definecolor{DsLogic}{HTML}{009E73}
\definecolor{DsPuzzle}{HTML}{A45F93}
\usepackage{array}
\usepackage{tabularx}
\usepackage{colortbl}
\usepackage{multirow}
\newcommand{\modelgroup}[2]{\multirow{-#1}{*}{\rotatebox[origin=c]{90}{\fontsize{6.5}{7}\selectfont#2-source}}}
\definecolor{ComparisonRow}{HTML}{E8F2FA}
\definecolor{HumanRow}{HTML}{EEEEEE}
\usepackage{algorithm}
\usepackage{algpseudocode}
\definecolor{PreprintLinkBlue}{rgb}{0.21,0.49,0.74}
\usepackage[pagebackref,breaklinks,colorlinks,allcolors=PreprintLinkBlue]{hyperref}

\title{Vis2State: Visual Reasoning through Executable State Abstractions}
\author{
Sheng Jin \quad Wentao Liu \quad Chen Qian\\[4pt]
SenseTime Research and Tetras.AI\\[4pt]
\{jinsheng, liuwentao, qianchen\}@tetras.ai
}

\hypersetup{pdftitle={Vis2State: Visual Reasoning through Executable State Abstractions},pdfauthor={Sheng Jin, Shaohua Li, Wentao Liu, Chen Qian},pdfsubject={Visual reasoning through executable state abstractions}}
\begin{document}

\maketitle

\begin{abstract}
Solving visual puzzles requires both accurate interpretation of visual evidence and reliable computation under task-specific rules. Embedding visual assumptions and task constraints directly in solving code makes the represented problem difficult to inspect and revise independently. We introduce Vis2State, a training-free agentic framework that abstracts visual puzzles into explicit, revisable problem states linking visual inspection and execution. Visual inspection guides state construction and refinement, while generated programs solve the represented problem. The model constructs both the state and the program without predefined puzzle schemas or solving procedures. Inspection and execution feedback allow it to revise the problem representation and solving procedure independently. Vis2State achieves the state-of-the-art performance on four benchmarks. Cross-model evaluation further demonstrates consistent gains. Ablations, error analyses, and fixed-program state interventions further support the effectiveness of the state-centered design.
The code will be made publicly available to facilitate future research.

\end{abstract}
\suppressfloats[t]

\begin{figure}[t]
\centering
\includegraphics[width=\columnwidth]{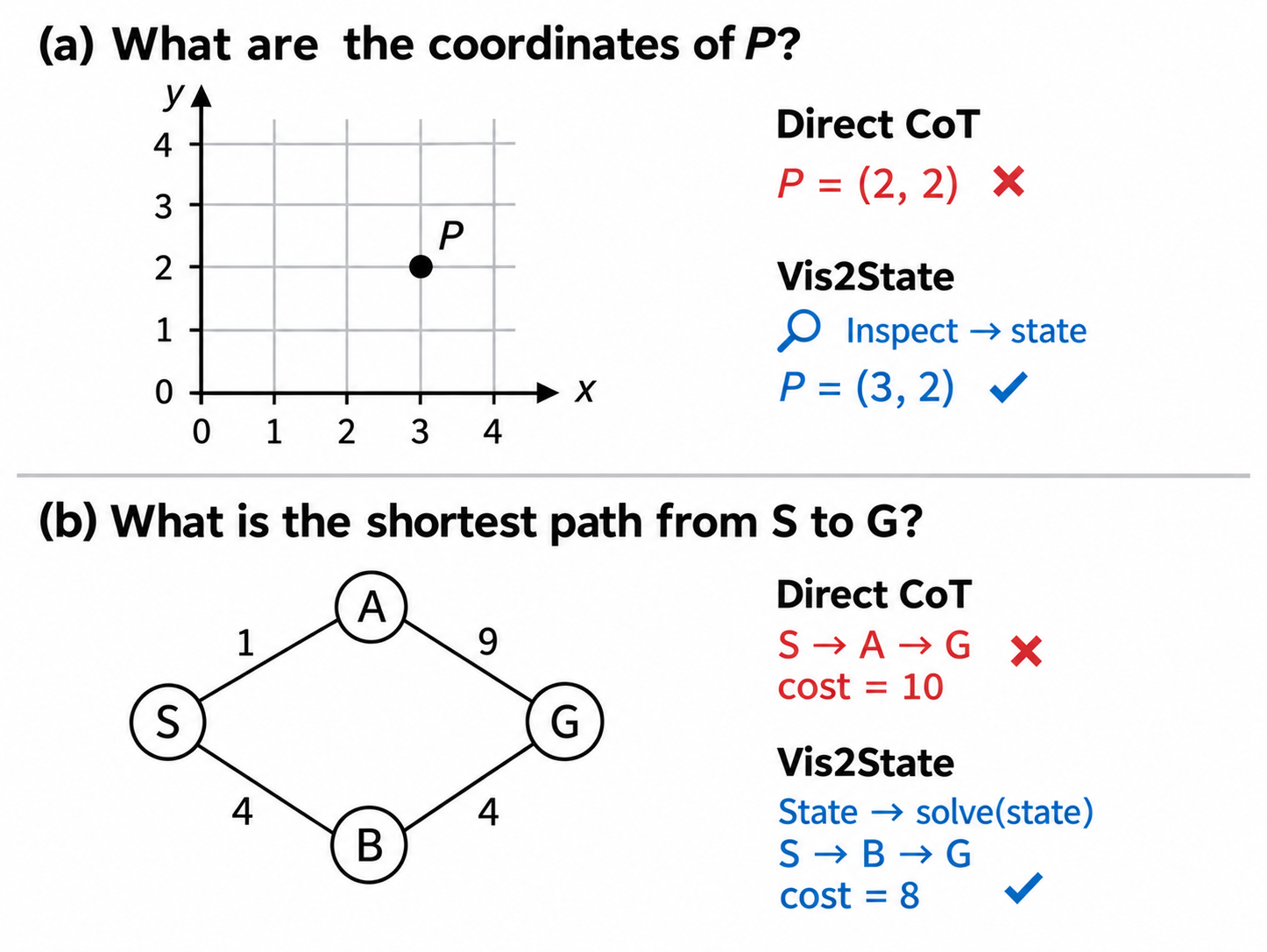}
\caption{\textbf{Vis2State turns visual evidence into explicit problem states for computational solving.} Two illustrative puzzles show how (a) inspection corrects a misread coordinate and (b) execution over explicit state finds the shortest path.}
\label{fig:principle-comparison}
\end{figure}

\section{Introduction}
Visual puzzles probe a model's ability to turn image observations into a solution under task-specific rules. Beyond recognizing objects, the model must infer their relations and task constraints to guide search or computation. Benchmarks spanning algorithmic puzzles, spatial reasoning, visual logic, and abstract patterns expose this connection between perception and reasoning~\cite{algopuzzlevqa,visualpuzzles,visulogic,puzzlevqa}. Their tightly coupled perceptual and computational demands make visual puzzles a revealing testbed for understanding where multimodal reasoning succeeds and where it breaks down.

Bridging visual perception and computation requires organizing visual evidence into a representation of the problem to be solved. The required information varies across tasks. Mazes require connectivity, color matching requires correspondences, and box pushing requires positions and legal transitions. Existing approaches provide complementary capabilities. Chain-of-thought (CoT) prompting elicits intermediate reasoning~\cite{cot,zeroshotcot}, visual search retrieves relevant details~\cite{vstar}, and visual programming connects perception to executable operations~\cite{visprog,viper}. These capabilities alone do not ensure that the represented problem and its solving procedure are both correct. A sound algorithm can fail on misinterpreted constraints, just as accurate observations can be undermined by incomplete search. When visual assumptions and task constraints are embedded directly in solving code, these two sources of error become entangled, making the problem representation difficult to inspect and revise independently.

Our central idea is that visual puzzles can be abstracted into explicit, revisable \emph{problem states}. We introduce \textbf{Vis2State}, a training-free agentic framework that uses these states as the interface between visual inspection and execution. Through inspection, the model constructs a task-dependent state encoding relevant objects, relations, constraints, and objectives, then generates a program to solve the represented problem. Neither the state schema nor the program is predefined. This separation allows the model to refine its interpretation of the puzzle independently of the procedure used to solve it.

The explicit state gives inspection and execution distinct but complementary roles. Inspection grounds the state in visual evidence by resolving uncertain objects, relations, or constraints. Execution takes the state and generated program as input, and returns computational results or errors. These observations guide the model's next step, which may involve further inspection, state revision, or program refinement. The feedback-driven process allows the model to alternate between establishing the problem and solving it, rather than follow a fixed inspection--execution sequence. Figure~\ref{fig:principle-comparison} illustrates both roles. Inspection corrects a misread coordinate in (a), while execution uses an explicit state to find the shortest path in (b).

Experiments on four visual reasoning benchmarks show that Vis2State achieves state-of-the-art accuracy, improving over Direct CoT by 8.1--15.5 percentage points. Cross-model evaluation confirms consistent gains, while ablations, error analyses, and fixed-program state interventions support the effectiveness of its state-centered design.

We make three contributions.
\begin{itemize}
\item \textbf{Explicit states as a reasoning interface.} We introduce Vis2State, a training-free agentic framework that abstracts visual puzzles into explicit, revisable problem states, separating the represented problem from its solving procedure.
\item \textbf{State-centered inspection and execution.} We develop an adaptive reasoning process that uses visual inspection to construct and refine states, and generated programs to solve the represented problems, with execution feedback guiding further revision.
\item \textbf{Strong performance and extensive analysis.} Vis2State achieves state-of-the-art accuracy on four benchmarks and consistent gains over Direct CoT baselines. Component ablations, error analyses, and fixed-program state interventions further substantiate the effectiveness of the state-centered design.
\end{itemize}

\section{Related Work}
\paragraph{Visual inspection and tool use.}
Visual tools support reasoning by acquiring and transforming task-relevant evidence. V*~\cite{vstar} guides visual search toward relevant image details. Visual Sketchpad~\cite{sketchpad} supports reasoning through intermediate sketches, while ReFocus~\cite{refocus} directs visual attention through highlighting and masking. Thyme~\cite{thyme} and CodeDance~\cite{codedance} use executable code for image manipulation and computation. To learn visual tool use, Pixel Reasoner~\cite{pixelreasoner} combines instruction tuning with curiosity-driven reinforcement learning, while DeepEyes~\cite{deepeyes} learns active perception through reinforcement learning. VTool-R1~\cite{vtoolr1} trains visual editing with outcome rewards, ToolsRL~\cite{toolsrl} introduces direct tool supervision, and VC-Tooler~\cite{vctooler} learns tool composition and adaptation to returned observations. CodeV~\cite{codev} evaluates evidence relevance and trains faithful tool use, while See2Think~\cite{seetothink} examines dependence on intermediate visual states. Vis2State directs inspection toward constructing and checking an explicit problem state that a generated program uses without access to the image. This interface makes the visual premises of computation explicit and revisable.

\paragraph{Program-based reasoning.}
Program-based methods make intermediate reasoning executable. Building on CoT~\cite{cot,zeroshotcot}, PAL~\cite{pal} and Program of Thoughts~\cite{pot} delegate computation to interpreters. Faithful CoT separates symbolic translation from deterministic solving~\cite{faithfulcot}, and Chain of Code combines execution with language-model emulation~\cite{chaincode}. Execution can also inform subsequent reasoning. ReAct interleaves reasoning with actions and observations~\cite{react}, while CodeAct uses executable code as its action space~\cite{codeact}. In vision, programs compose visual modules~\cite{iep,visprog,viper}, and explicit scene representations support symbolic reasoning~\cite{nsvqa,nscl}. Vis2State centers this process on a revisable \emph{problem state} supplied to a generated program that runs without image access. It represents task-dependent objects, relations, constraints, and objectives separately from the program, allowing the problem representation and solving procedure to be revised independently.

\begin{figure*}[t]
\centering
\includegraphics[width=\textwidth,trim={0 110bp 0 0},clip]{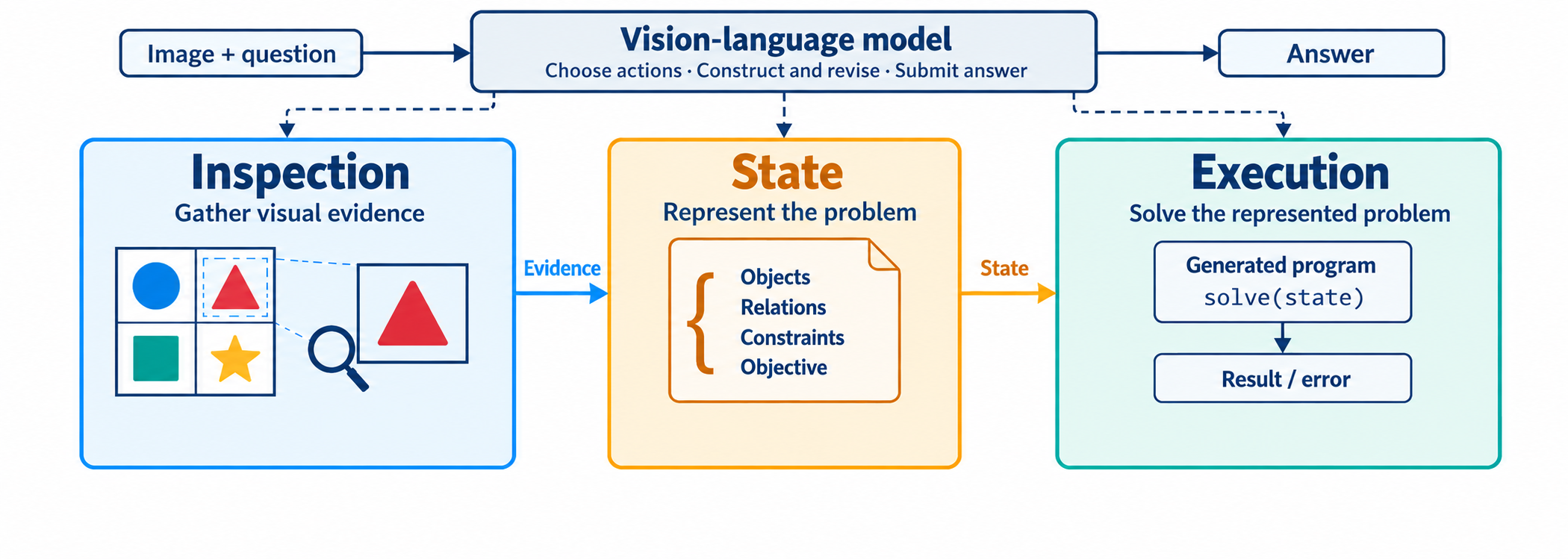}
\caption{\textbf{Inspection, State, and Execution in Vis2State.}
\emph{Inspection} gathers visual evidence to construct the \emph{State}, and \emph{Execution} solves the represented problem. The model refines the state or program through feedback before submitting an answer.}
\label{fig:vis2state-overview}
\end{figure*}

\section{Vis2State}
\label{sec:method}
Vis2State connects visual inspection and computational solving through an explicit problem state (Figure~\ref{fig:vis2state-overview}). Inspection provides evidence to construct and refine the state, while execution applies a generated program to solve the represented problem. A vision-language model $M_\theta$ coordinates these complementary processes, using accumulated feedback to determine what to inspect, revise, or compute next. Given an image $x$ and question $q$, the model selects its next action as
\begin{equation}
a_t=M_\theta(x,q,H_{t-1};\mathcal{P}),\qquad H_0=[],
\label{eq:action-policy}
\end{equation}
where $\mathcal{P}$ is the shared prompt and $H_{t-1}$ records prior actions and observations. Each action $a_t$ inspects the image, executes a program, or submits an answer. Subject to call and time budgets, the model can revisit inspection or execution as needed, rather than follow a fixed sequence.
All datasets and models share the same prompt and tool instructions. Details appear in the supplementary material.

\subsection{State Abstraction}
The problem state encodes task-relevant information from the image and question independently of the solving procedure. For an execution action at step $t$, we represent the problem state as
\begin{equation}
s_t=\{k_i:v_i\}_{i=1}^{K_t},
\label{eq:state-program}
\end{equation}
where $s_t$ is a JSON object with $K_t$ key--value fields. The model constructs the fields $k_i$ and their values $v_i$ to represent task-relevant objects, relations, constraints, and objectives, without predefined task- or model-specific schemas.

Keeping this representation separate from the program allows the model to revise its interpretation of the puzzle independently of the solving procedure. In box pushing, for example, the state records walls, player, box, and goal positions, while the program implements legal transitions and search. A misread wall requires correcting the state, whereas an incorrect transition implementation requires revising the program. The model can update either component in subsequent actions, with prior observations and results retained in the interaction history. Inspection and answer submission do not require a state. Recorded state examples appear in the supplement.

\begin{table}[tbp]
\centering\small
\caption{Visual inspection tools, with interaction design informed by V*~\cite{vstar} and Visual Sketchpad~\cite{sketchpad}.}
\label{tab:tools}
\begin{tabular}{@{}p{.38\linewidth}p{\dimexpr.62\linewidth-2\tabcolsep\relax}@{}}
\toprule
Tool & Output\\
\midrule
\texttt{crop\_region} & Cropped image with coordinates.\\
\texttt{pixel\_probe} & RGB values at queried points and regional color statistics.\\
\texttt{edge\_view} & Image of local RGB differences.\\
\texttt{color\_extent} & Bounding box and count of matched pixels, with an annotated image.\\
\texttt{color\_components} & Bounding boxes and sizes of connected color components, with an annotated image.\\
\bottomrule
\end{tabular}
\end{table}

\subsection{Visual Inspection}
Visual inspection grounds the problem state in image evidence. The model requests targeted views or measurements to guide state construction and resolve uncertainty in the current representation. At step $t$, it specifies the tools and their arguments in a request list $R_t$, yielding
\begin{equation}
a_t=\operatorname{Inspect}(x,R_t)\;\longrightarrow\;o_t,
\label{eq:inspection}
\end{equation}
where $o_t$ denotes the observation returned after executing action $a_t$. The tools in Table~\ref{tab:tools} perform deterministic image operations, while the model interprets their outputs in the context of the task. For example, a connected color component identifies a pixel group, but the model must determine whether that group corresponds to a wall, goal, or another task entity. These observations ground task-relevant attributes and relations in visual evidence.

The model adaptively selects inspection calls from accumulated observations, either before execution or when computational feedback raises uncertainty about the state.

\subsection{Execution and Feedback}
For execution, the model supplies a state--program pair $(s_t,f_t)$, where $f_t$ defines \texttt{solve(state)}. Execution applies $f_t$ to $s_t$ and returns an observation,
\begin{equation}
a_t=\operatorname{Exec}(s_t,f_t)\;\longrightarrow\;o_t.
\label{eq:execution}
\end{equation}
The program takes the supplied state as input, without access to the image, and returns a computational result or error. After each nonterminal interaction, the action and observation are appended to the history,
\begin{equation}
H_t=H_{t-1}\mathbin{\|}[(a_t,o_t)],
\label{eq:history-update}
\end{equation}
where $\|$ denotes sequence concatenation. Using this history, the model can inspect uncertain visual details, revise the state or program, or submit an answer. Execution feedback thus supports iterative refinement of both the problem representation and the solving procedure.

\begin{table*}[!t]
\centering\footnotesize
\setlength{\tabcolsep}{2.2pt}
\caption{\textbf{Accuracy (\%) across 18 task families on AlgoPuzzleVQA~\cite{algopuzzlevqa}.} Bold indicates the best model result in each column.}
\label{tab:benchmark-algopuzzlevqa}
\begin{tabularx}{\textwidth}{@{}c@{\hspace{4pt}}l*{10}{>{\raggedleft\arraybackslash}X}@{}}
\toprule
 & Method & Overall & Board & Calendar & Chain & Checker & Clock & Colour & Map & Maze & Move box \\
\midrule
 & GPT-4V~\cite{gpt4vcard} & 30.3 & 48.0 & 57.0 & 21.0 & 33.0 & 25.0 & 27.0 & 34.0 & 30.0 & 29.0 \\
 & Claude 3 Opus~\cite{claude3} & 30.9 & 48.0 & 39.0 & 23.0 & 35.0 & 37.0 & 16.0 & 33.0 & 23.0 & 19.0 \\
 & GPT-5.2~\cite{gpt52card} & 54.6 & 49.0 & 70.0 & 23.0 & 56.0 & 76.0 & 38.0 & 51.0 & 38.0 & 39.0 \\
 & o3 (high)~\cite{o3o4card} & 62.3 & 60.0 & 96.0 & 75.0 & 56.0 & 98.0 & 47.0 & 5.0 & 39.0 & 52.0 \\
 & o4-mini (high)~\cite{o3o4card} & 65.4 & 52.0 & 99.0 & 77.0 & 97.0 & 91.0 & 47.0 & 3.0 & 42.0 & 50.0 \\
\modelgroup{6}{Closed} & Gemini 3 Pro Preview~\cite{gemini3card} & 81.3 & 93.0 & 99.0 & 61.0 & 94.0 & 98.0 & 43.0 & 61.0 & 49.0 & 87.0 \\
\midrule
 & Qwen2-VL-7B~\cite{qwen2vl} & 28.2 & 45.0 & 28.0 & 31.0 & 33.0 & 26.0 & 20.0 & 32.0 & 25.0 & 27.0 \\
 & Qwen3-VL-8B-Instruct~\cite{qwen3vlreport} & 45.0 & 45.0 & 70.0 & 26.0 & 64.0 & 55.0 & 30.0 & 44.0 & 26.0 & 27.0 \\
\rowcolor{ComparisonRow}
\cellcolor{white} & DeepSeek V4.1 Flash~\cite{deepseek41} & 82.3 & 91.0 & 97.0 & 97.0 & \textbf{100.0} & 96.0 & 49.0 & 66.0 & 41.0 & 57.0 \\
\rowcolor{ComparisonRow}
\cellcolor{white}\modelgroup{4}{Open} & DeepSeek V4.1 Flash + Vis2State (Ours) & \textbf{97.8} & \textbf{100.0} & \textbf{100.0} & \textbf{100.0} & \textbf{100.0} & \textbf{100.0} & \textbf{100.0} & \textbf{81.0} & \textbf{100.0} & \textbf{100.0} \\
\bottomrule
\end{tabularx}
\par\vspace{5pt}
\begin{tabularx}{\textwidth}{@{}c@{\hspace{4pt}}l*{10}{>{\raggedleft\arraybackslash}X}@{}}
\toprule
 & Method & Overall & Queens & Number & Fruits & Rubik & Think dot & Hanoi & Water & Wheel & Wood \\
\midrule
 & GPT-4V~\cite{gpt4vcard} & 30.3 & 14.0 & 41.0 & 27.0 & 48.0 & 31.0 & 17.0 & 17.0 & 25.0 & 21.0 \\
 & Claude 3 Opus~\cite{claude3} & 30.9 & 15.0 & 43.0 & 40.0 & 39.0 & 53.0 & 22.0 & 37.0 & 19.0 & 16.0 \\
 & GPT-5.2~\cite{gpt52card} & 54.6 & 98.0 & 83.0 & 74.0 & 72.0 & 56.0 & 42.0 & 48.0 & 40.0 & 30.0 \\
 & o3 (high)~\cite{o3o4card} & 62.3 & 35.0 & 30.0 & 84.0 & 85.0 & 69.0 & 79.0 & 88.0 & 75.0 & 48.0 \\
 & o4-mini (high)~\cite{o3o4card} & 65.4 & 56.0 & 29.0 & 97.0 & 92.0 & 74.0 & 85.0 & 82.0 & 69.0 & 36.0 \\
\modelgroup{6}{Closed} & Gemini 3 Pro Preview~\cite{gemini3card} & 81.3 & 96.0 & 94.0 & \textbf{100.0} & 88.0 & 74.0 & 83.0 & 84.0 & 81.0 & 78.0 \\
\midrule
 & Qwen2-VL-7B~\cite{qwen2vl} & 28.2 & 36.0 & 34.0 & 30.0 & 37.0 & 27.0 & 23.0 & 18.0 & 31.0 & 23.0 \\
 & Qwen3-VL-8B-Instruct~\cite{qwen3vlreport} & 45.0 & 36.0 & 76.0 & 71.0 & 50.0 & 56.0 & 20.0 & 50.0 & 50.0 & 14.0 \\
\rowcolor{ComparisonRow}
\cellcolor{white} & DeepSeek V4.1 Flash~\cite{deepseek41} & 82.3 & 76.0 & \textbf{100.0} & \textbf{100.0} & 97.0 & 81.0 & 86.0 & 95.0 & 88.0 & 64.0 \\
\rowcolor{ComparisonRow}
\cellcolor{white}\modelgroup{4}{Open} & DeepSeek V4.1 Flash + Vis2State (Ours) & \textbf{97.8} & \textbf{100.0} & \textbf{100.0} & \textbf{100.0} & \textbf{100.0} & \textbf{87.0} & \textbf{100.0} & \textbf{100.0} & \textbf{96.0} & \textbf{96.0} \\
\bottomrule
\end{tabularx}
\end{table*}

\begin{table*}[!t]
\centering\footnotesize
\setlength{\tabcolsep}{2.2pt}
\caption{\textbf{Accuracy (\%) by reasoning category on VisualPuzzles~\cite{visualpuzzles}.} Bold indicates the best model result in each column.}
\label{tab:benchmark-visualpuzzles}
\begin{tabularx}{\textwidth}{@{}c@{\hspace{4pt}}l*{6}{>{\raggedleft\arraybackslash}X}@{}}
\toprule
 & Method & Overall & Algorithmic & Analogical & Deductive & Inductive & Spatial \\
\midrule
\rowcolor{HumanRow}
 & Human (50th Percentile) & 75.0 & 88.0 & 66.0 & 80.0 & 50.0 & 90.0 \\
\midrule
 & Claude 3.7 Sonnet~\cite{claude37} & 48.3 & 64.5 & 48.3 & 65.0 & 26.8 & 37.4 \\
 & Claude Opus 4.1 (thinking)~\cite{claude41} & 48.8 & -- & -- & -- & -- & -- \\
 & Gemini 2.5 Pro~\cite{gemini25} & 49.5 & 60.0 & 64.0 & 60.0 & 29.7 & 36.4 \\
 & o3~\cite{o3o4card} & 54.0 & 64.5 & 68.3 & 69.5 & 27.3 & 42.7 \\
 & o4-mini~\cite{o3o4card} & 57.0 & 65.3 & 68.7 & 75.5 & 33.0 & 45.5 \\
 & GPT-5 (high)~\cite{gpt5card} & 57.3 & -- & -- & -- & -- & -- \\
\modelgroup{7}{Closed} & Gemini 3.1 Pro~\cite{gemini31card} & 73.5 & -- & -- & -- & -- & -- \\
\midrule
 & Qwen3-VL-8B-Instruct~\cite{qwen3vlreport} & 36.2 & -- & -- & -- & -- & -- \\
 & Qwen2.5-VL-72B-Instruct~\cite{qwen25vl} & 42.3 & 53.4 & 46.9 & 58.0 & 25.8 & 29.5 \\
 & Qwen3-VL-235B-A22B (instruct)~\cite{qwen3vlreport} & 54.7 & -- & -- & -- & -- & -- \\
 & Qwen3-VL-235B-A22B (thinking)~\cite{qwen3vlreport} & 57.2 & -- & -- & -- & -- & -- \\
\rowcolor{ComparisonRow}
\cellcolor{white} & DeepSeek V4.1 Flash~\cite{deepseek41} & 61.6 & 77.1 & 63.0 & 83.5 & 38.8 & 47.6 \\
\rowcolor{ComparisonRow}
\cellcolor{white}\modelgroup{6}{Open} & DeepSeek V4.1 Flash + Vis2State (Ours) & \textbf{75.0} & \textbf{92.0} & \textbf{76.3} & \textbf{92.5} & \textbf{47.4} & \textbf{66.4} \\
\bottomrule
\end{tabularx}
\end{table*}

\begin{table*}[!t]
\centering\footnotesize
\setlength{\tabcolsep}{2.2pt}
\caption{\textbf{Accuracy (\%) across six reasoning categories on VisuLogic~\cite{visulogic}.} Bold indicates the best model result in each column.}
\label{tab:benchmark-visulogic}
\begin{tabularx}{\textwidth}{@{}c@{\hspace{4pt}}l*{7}{>{\raggedleft\arraybackslash}X}@{}}
\toprule
 & Method & Overall & Quantity & Spatiality & Position & Attribute & Style & Other \\
\midrule
\rowcolor{HumanRow}
 & Human & 51.4 & 45.3 & 52.7 & 71.1 & 50.0 & 47.5 & 44.2 \\
\midrule
 & GPT-4o~\cite{gpt4ocard} & 26.3 & 28.6 & 24.7 & 27.2 & 26.8 & 20.0 & 25.9 \\
 & Claude Opus 4.1 (thinking)~\cite{claude41} & 27.9 & -- & -- & -- & -- & -- & -- \\
 & Gemini 2.0 Pro~\cite{gemini20} & 28.0 & 29.7 & 24.2 & 27.9 & 30.5 & 22.2 & 33.3 \\
 & GPT-5 (high)~\cite{gpt5card} & 28.5 & -- & -- & -- & -- & -- & -- \\
 & o3~\cite{o3o4card} & 29.5 & 24.2 & 31.0 & 34.6 & 27.0 & 25.0 & 41.9 \\
 & Gemini 2.5 Pro (thinking)~\cite{gemini25} & 31.6 & -- & -- & -- & -- & -- & -- \\
 & Claude Opus 4.7~\cite{claude47} & 32.6 & -- & -- & -- & -- & -- & -- \\
 & Doubao 1.6 Vision~\cite{doubao16vision} & 34.9 & 32.9 & 34.6 & 39.0 & 31.7 & 30.0 & 43.5 \\
 & GPT-5.5~\cite{gpt55card} & 43.5 & -- & -- & -- & -- & -- & -- \\
\modelgroup{10}{Closed} & Gemini 3.1 Pro~\cite{gemini31card} & 44.8 & -- & -- & -- & -- & -- & -- \\
\midrule
 & Qwen2.5-VL-72B~\cite{qwen25vl} & 26.2 & 25.2 & 23.8 & 27.2 & 25.6 & 25.6 & 34.3 \\
 & InternVL3-78B~\cite{internvl3} & 27.7 & 27.7 & 26.1 & 31.6 & 26.3 & 21.3 & 32.3 \\
 & Qwen3-VL-235B-A22B (instruct)~\cite{qwen3vlreport} & 29.9 & -- & -- & -- & -- & -- & -- \\
 & Qwen3-VL-235B-A22B (thinking)~\cite{qwen3vlreport} & 34.4 & -- & -- & -- & -- & -- & -- \\
\rowcolor{ComparisonRow}
\cellcolor{white} & DeepSeek V4.1 Flash~\cite{deepseek41} & 35.5 & 34.0 & 37.7 & 38.2 & 32.9 & 23.3 & 44.4 \\
\rowcolor{ComparisonRow}
\cellcolor{white}\modelgroup{6}{Open} & DeepSeek V4.1 Flash + Vis2State (Ours) & \textbf{49.1} & \textbf{47.9} & \textbf{39.8} & \textbf{68.4} & \textbf{45.1} & \textbf{46.7} & \textbf{53.7} \\
\bottomrule
\end{tabularx}
\end{table*}

\begin{table*}[!t]
\centering\scriptsize
\setlength{\tabcolsep}{1.5pt}
\renewcommand{\arraystretch}{1.5}
\caption{\textbf{Accuracy (\%) on single- and dual-concept puzzles in PuzzleVQA~\cite{puzzlevqa}.} N/C/S/Z denote numbers/colors/shapes/size. Bold indicates the best model result in each column.}
\label{tab:benchmark-puzzlevqa}
\begin{tabularx}{\textwidth}{@{}c@{\hspace{4pt}}l*{13}{>{\raggedleft\arraybackslash}X}@{}}
\toprule
 & & & \multicolumn{5}{c}{Single-concept} & \multicolumn{7}{c}{Dual-concept} \\
\cmidrule(lr){4-8}\cmidrule(lr){9-15}
 & Method & Overall & N & C & Z & S & Avg. & N$+$S & C$+$N & N$+$Z & C$+$S & Z$+$S & C$+$Z & Avg. \\
\midrule
 & Claude 3 Opus~\cite{claude3} & 42.0 & 47.0 & 32.5 & 33.5 & 44.5 & 39.4 & 48.0 & 54.5 & 34.0 & 42.5 & 33.0 & 50.0 & 43.7 \\
 & GPT-4V~\cite{gpt4vcard} & 45.9 & 67.5 & 42.0 & 35.0 & 41.0 & 46.4 & 52.5 & 56.0 & 31.0 & 47.0 & 31.5 & 55.0 & 45.5 \\
 & GPT-5.2~\cite{gpt52card} & 67.3 & 69.5 & 98.0 & 66.0 & 71.5 & 76.3 & 75.0 & 68.5 & 37.0 & 61.0 & 68.5 & 58.5 & 61.4 \\
 & o3 (high)~\cite{o3o4card} & 84.0 & \textbf{99.5} & 99.0 & 87.5 & 74.0 & 90.0 & 92.0 & 97.5 & 45.0 & 92.0 & 97.0 & 57.0 & 80.1 \\
 & o4-mini (high)~\cite{o3o4card} & 85.7 & 99.0 & \textbf{99.5} & 86.0 & 70.5 & 88.8 & 96.0 & 99.5 & 62.0 & 86.5 & 85.5 & 72.5 & 83.7 \\
\modelgroup{6}{Closed} & Gemini 3 Pro Preview~\cite{gemini3card} & 86.3 & 99.0 & 94.0 & 83.5 & \textbf{77.0} & 88.4 & \textbf{99.0} & 92.5 & 73.5 & 84.5 & 88.5 & 71.5 & 84.9 \\
\midrule
 & Qwen3-VL-8B-Instruct~\cite{qwen3vlreport} & 59.9 & 80.0 & 85.5 & 34.5 & 74.0 & 68.5 & 71.5 & 70.5 & 35.5 & 39.0 & 49.5 & 58.5 & 54.1 \\
\rowcolor{ComparisonRow}
\cellcolor{white} & DeepSeek V4.1 Flash~\cite{deepseek41} & 85.2 & 93.0 & 95.5 & 83.0 & 70.5 & 85.5 & 92.0 & 98.5 & 61.5 & 92.5 & 89.5 & 76.0 & 85.0 \\
\rowcolor{ComparisonRow}
\cellcolor{white}\modelgroup{3}{Open} & DeepSeek V4.1 Flash + Vis2State (Ours) & \textbf{93.3} & 96.0 & \textbf{99.5} & \textbf{100.0} & 72.0 & \textbf{91.9} & 94.5 & \textbf{100.0} & \textbf{86.0} & \textbf{97.5} & \textbf{98.5} & \textbf{88.5} & \textbf{94.2} \\
\bottomrule
\end{tabularx}
\end{table*}

\section{Experiments and Analysis}

\subsection{Experimental protocol}
\label{sec:datasets}
\paragraph{Datasets.}
We evaluate 5,968 multiple-choice questions across four benchmarks. \textbf{AlgoPuzzleVQA}~\cite{algopuzzlevqa} contains 1,800 questions in 18 algorithmic puzzle families, including mazes, box pushing, map coloring, and the Tower of Hanoi, which require rule-constrained search and computation. \textbf{VisualPuzzles}~\cite{visualpuzzles} contains 1,168 questions spanning algorithmic, analogical, deductive, inductive, and spatial reasoning, with little specialized knowledge required. \textbf{VisuLogic}~\cite{visulogic} contains 1,000 vision-centric logic questions across Quantity, Spatiality, Position, Attribute, Style, and Other, covering counting, folding, rotation, and graphical composition to test visual attributes and relations. \textbf{PuzzleVQA}~\cite{puzzlevqa} contains 2,000 abstract-pattern questions from 20 templates, covering numbers, colors, shapes, and size through four single-concept and six dual-concept categories requiring visual pattern induction.

\paragraph{Evaluation protocol.}
\label{sec:scoring}
We report accuracy using each benchmark's released answer-extraction and scoring procedures. Direct CoT follows the official CoT prompt for each benchmark~\cite{algopuzzlevqa,visualpuzzles,visulogic,puzzlevqa}. Following VisualPuzzles~\cite{visualpuzzles}, we uniformly sample a valid option with a fixed seed when no final answer is produced. For VisuLogic evaluation, we use DeepSeek V4.1 Flash in place of the released evaluator's GPT-4o-mini~\cite{gpt4omini} because of model availability.

\paragraph{Implementations.}
\label{sec:implementation}
DeepSeek-V4.1-Flash-260910~\cite{deepseek41} is used for the main experiments, and GLM-5.3-Flash-260828~\cite{glm5} for cross-model evaluation. Models receive the image and question, including answer choices, without reference answers or explanations. In our experiments, thinking is enabled with a maximum output length of 65,536 tokens per model call. Truncated responses return step-error feedback and allow continuation within the remaining budgets. The supplement provides prompt and runtime details.

\begin{figure*}[!t]
\centering
\includegraphics[width=\textwidth]{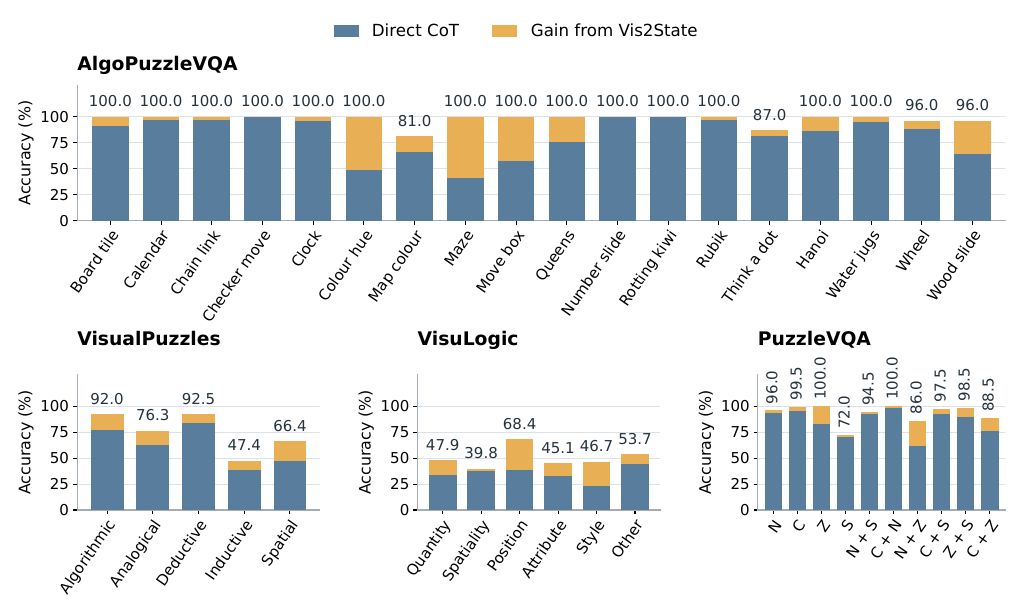}
\caption{\textbf{Category-level accuracy and gains over Direct CoT.} Blue bars show Direct CoT accuracy. Orange segments show the accuracy gain from Vis2State. Each full bar and its numeric label give Vis2State accuracy (\%). In PuzzleVQA, N/C/S/Z denote numbers/colors/shapes/size. Vis2State improves 36 of 39 categories and ties the remaining three at 100.0\%.}
\label{fig:category-gains}
\end{figure*}

\subsection{Benchmark comparison}
Vis2State achieves state-of-the-art overall accuracy among the compared methods on all four benchmarks (Tables~\ref{tab:benchmark-algopuzzlevqa}--\ref{tab:benchmark-puzzlevqa}). With DeepSeek V4.1 Flash, it outperforms Direct CoT by 8.1--15.5 percentage points, demonstrating substantial gains over the same model without tools.

The improvements extend to 36 of 39 categories, with the remaining three tied at 100.0\% (Figure~\ref{fig:category-gains}). On AlgoPuzzleVQA, Vis2State achieves 100.0\% accuracy in 14 of 18 puzzle families, compared with three for Direct CoT. Gains span every category of VisualPuzzles and VisuLogic, with the largest improvements in Spatial reasoning and Position, respectively. Explicit states make spatial configurations and positional relations available for inspection and computation, providing a way to check visual interpretations and reason over the represented problem. Gains are smaller on VisuLogic Spatiality and PuzzleVQA Shapes, suggesting that some perceptual and relational ambiguities remain difficult to resolve.

\begin{table}[tb]
\centering\footnotesize
\setlength{\tabcolsep}{3pt}
\caption{Paired comparison with Direct CoT. \# Wins count questions answered correctly only by Vis2State. \# Losses count those answered correctly only by Direct CoT. Intervals are paired 95.0\% bootstrap confidence intervals for the accuracy gain.}
\label{tab:paired-comparison}
\begin{tabular*}{\linewidth}{@{\extracolsep{\fill}}lrrc@{}}
\toprule
Benchmark & \# Wins & \# Losses & Gain [95.0\% CI] \\
\midrule
AlgoPuzzleVQA & 295 & 16 & 15.5 [13.7, 17.3] \\
VisualPuzzles & 237 & 80 & 13.4 [10.5, 16.4] \\
VisuLogic & 238 & 102 & 13.6 [10.1, 17.1] \\
PuzzleVQA & 208 & 47 & 8.1 [6.6, 9.6] \\
\bottomrule
\end{tabular*}
\end{table}

Paired comparisons show that Vis2State corrects substantially more Direct CoT errors than it introduces on every benchmark (Table~\ref{tab:paired-comparison}). The gains are significant under paired tests with Holm correction across the four benchmarks ($p<2\times10^{-12}$), supporting the consistency of the overall improvements.

\begin{table*}[!t]
\centering\small
\setlength{\tabcolsep}{3pt}
\caption{\textbf{Ablation study.} We compare visual inspection, code execution, and their combination using DeepSeek V4.1 Flash. Vis2State connects inspection and execution through an explicit state--program interface and achieves the highest accuracy on all four benchmarks. Checkmarks indicate the components included in each configuration. Results are accuracy (\%), with the best score in each column in bold.}
\label{tab:main}
\begin{tabular*}{\textwidth}{@{\extracolsep{\fill}}l|ccc|cccc@{}}
\toprule
\multirow{2}{*}{Configuration} & \multirow{2}{1.35cm}{\centering Visual\newline Inspection} & \multirow{2}{1.15cm}{\centering Explicit\newline State} & \multirow{2}{1.35cm}{\centering Code\newline Execution} & \multirow{2}{*}{AlgoPuzzleVQA} & \multirow{2}{*}{VisualPuzzles} & \multirow{2}{*}{VisuLogic} & \multirow{2}{*}{PuzzleVQA}\\
 & & & & & & & \\
\midrule
Direct CoT & $\times$ & $\times$ & $\times$ & 82.3 & 61.6 & 35.5 & 85.2\\
Inspection only & $\checkmark$ & $\times$ & $\times$ & 92.1 & 72.0 & 43.0 & 92.1\\
Execution only & $\times$ & $\times$ & $\checkmark$ & 90.0 & 64.0 & 37.7 & 89.5\\
Inspection + Execution & $\checkmark$ & $\times$ & $\checkmark$ & 96.9 & 73.5 & 43.1 & 91.7\\
Vis2State & $\checkmark$ & $\checkmark$ & $\checkmark$ & \textbf{97.8} & \textbf{75.0} & \textbf{49.1} & \textbf{93.3}\\
\bottomrule
\end{tabular*}
\end{table*}

\begin{figure*}[!t]
\centering
\includegraphics[width=\textwidth]{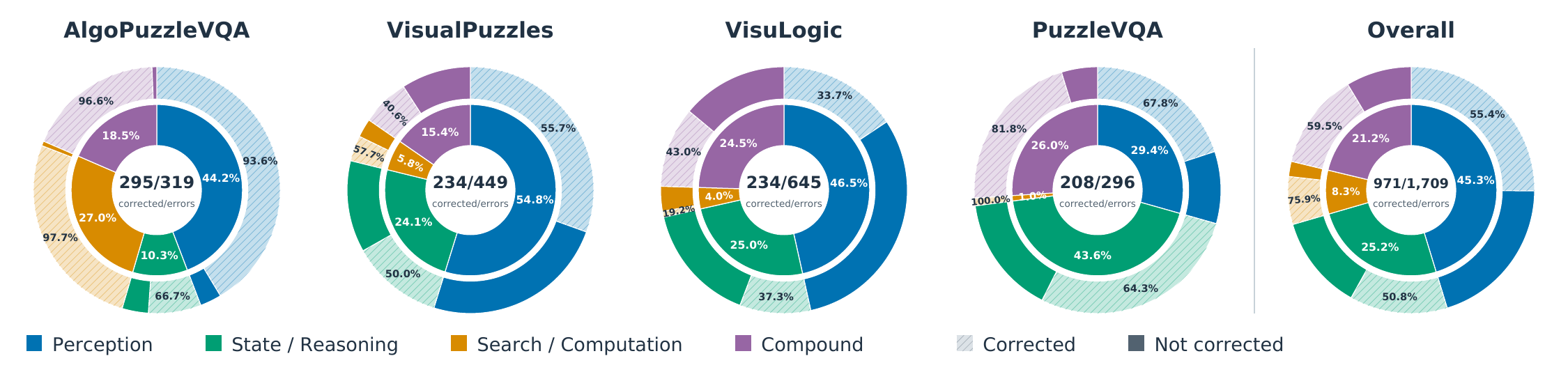}
\caption{\textbf{Direct CoT error profiles and corrections by Vis2State.}
Inner rings show Direct CoT error-group proportions. Outer rings show cases corrected by Vis2State (light, hatched) or not corrected (dark, solid), with within-group correction rates. Centers report corrected/total errors. State / Reasoning combines representation and rule errors; Search / Computation combines search, state updates, arithmetic, and answer comparison. Compound includes cross-group, answer-extraction, random-parser, and unattributed errors.}
\label{fig:direct-diagnosis}
\end{figure*}

\subsection{Component ablations}
We examine how visual inspection, program execution, and explicit problem states contribute to accuracy using five configurations of the same DeepSeek model (Table~\ref{tab:main}). \textbf{Direct CoT} answers without tools using the benchmark's official CoT prompt. \textbf{Inspection only} adds the five visual inspection tools, while \textbf{Execution only} provides Python execution and feedback. \textbf{Inspection + Execution} combines both capabilities without requiring an explicit state. \textbf{Vis2State} further introduces state-construction guidance and the \texttt{solve(state)} interface.

Inspection only improves over Direct CoT on all four benchmarks. Vis2State consistently outperforms Inspection + Execution by 0.8--6.0 percentage points, with the largest gain on VisuLogic. These gains show the benefit of using explicit, revisable states to coordinate inspection and execution.

\subsection{Error patterns and corrections}
\label{sec:direct-diagnosis}
We examine where Direct CoT fails and which failures Vis2State resolves on the same 1,709 diagnostic questions (Figure~\ref{fig:direct-diagnosis}), using the annotation protocol in the supplement. The dominant error types differ across benchmarks. Perception is the largest group on AlgoPuzzleVQA, VisualPuzzles, and VisuLogic, accounting for 44.2\%, 54.8\%, and 46.5\% of errors, respectively. On PuzzleVQA, State / Reasoning accounts for 43.6\%, highlighting difficulties in identifying the relations underlying visual patterns. These profiles distinguish failures in extracting visual evidence from failures in formulating and solving the represented problem.

Vis2State correctly answers 971 of these questions through valid submissions, excluding random completions. Correction rates vary across error groups. Search / Computation has the highest aggregate rate at 75.9\%, largely reflecting AlgoPuzzleVQA, where 97.7\% of cases are corrected. Perception yields the largest number of corrections, but its correction rate is 55.4\%; State / Reasoning remains lower at 50.8\%. This pattern is consistent with execution supporting systematic search and computation once a suitable state is available. 
The lower correction rates for Perception and State / Reasoning suggest that extracting visual information and inferring the relevant rules remain challenging for Vis2State.

\begin{figure*}[!t]
\centering
\includegraphics[width=\textwidth]{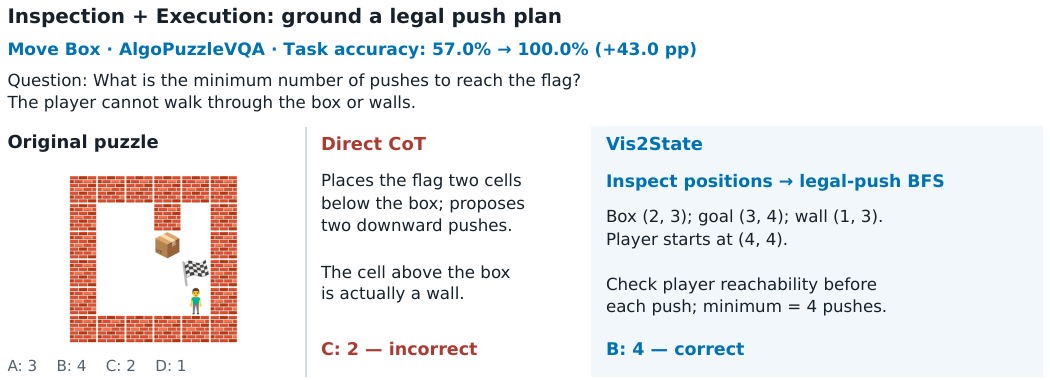}
\caption{\textbf{Case study on Move Box.}
Direct CoT misreads the goal position and overlooks a wall, producing an incorrect two-push solution. Vis2State explicitly represents the puzzle state and searches legal push sequences, finding the minimum of four pushes.}
\label{fig:current-direct-case}
\end{figure*}

\subsection{Counterfactual interventions on state}
\label{sec:actual-state-use}
We select 200 questions each from AlgoPuzzleVQA and PuzzleVQA with correct original-state outputs; original-state accuracy is 100.0\% by construction. Holding programs and answer options fixed, we test original states, answer-changing counterfactuals, and answer-preserving controls. Independent reference programs verify that each counterfactual maps to a different existing answer option and that each control preserves the original answer. Selection does not depend on intervention success: all incorrect intervention outputs remain in the evaluation. The supplement details the protocol and subset composition.

\begin{table}[tb]
\centering\small
\setlength{\tabcolsep}{3pt}
\caption{\textbf{State interventions.}
Correct counts and accuracies on 200 questions per dataset.}
\label{tab:state-counterfactual}
\begin{tabular*}{\linewidth}{@{\extracolsep{\fill}}lrr@{}}
\toprule
State condition & AlgoPuzzleVQA & PuzzleVQA \\
\midrule
Original state & 200 (100.0\%) & 200 (100.0\%) \\
Counterfactual & 196 (98.0\%) & 187 (93.5\%) \\
Answer-preserving control & 200 (100.0\%) & 197 (98.5\%) \\
\bottomrule
\end{tabular*}
\end{table}

Table~\ref{tab:state-counterfactual} shows that the saved programs correctly answer 98.0\% of counterfactuals on AlgoPuzzleVQA and 93.5\% on PuzzleVQA. Counterfactual accuracy falls below 100\% because some programs make reasoning or answer-mapping errors under the modified states. Under answer-preserving controls, accuracy remains at 100.0\% and 98.5\%, respectively. On these subsets, programs generally adapt their answers to state changes while preserving correctness under answer-preserving edits.

\subsection{Cross-model evaluation}
\label{sec:cross-model}

\begin{table}[tbp]
\centering\small
\setlength{\tabcolsep}{4pt}
\caption{\textbf{Cross-model evaluation.} Accuracy (\%) with GLM-5.3-Flash on AlgoPuzzleVQA (AP), VisualPuzzles (VP), VisuLogic (VL), and PuzzleVQA (PQ).}
\label{tab:cross-model}
\begin{tabular*}{\linewidth}{@{\extracolsep{\fill}}lrrrr@{}}
\toprule
Method & AP & VP & VL & PQ \\
\midrule
Direct CoT (GLM-5.3-Flash) & 82.3 & 67.2 & 36.2 & 89.4 \\
Vis2State (GLM-5.3-Flash) & \textbf{94.5} & \textbf{75.1} & \textbf{44.2} & \textbf{96.3} \\
\midrule
Gain (pp) & +12.2 & +7.9 & +8.0 & +6.9 \\
\bottomrule
\end{tabular*}
\end{table}

To assess whether Vis2State generalizes across models, we replace DeepSeek V4.1 Flash with GLM-5.3-Flash-260828~\cite{glm5}, keeping the prompt, tool instructions, and state interface unchanged. Table~\ref{tab:cross-model} shows consistent improvements over Direct CoT with the same model, with gains of 6.9--12.2 percentage points across all four benchmarks. These results extend the benefits of Vis2State to a second model without model-specific adaptation.

The benchmark difficulty pattern also remains consistent across the two models. GLM-based Vis2State achieves its highest accuracies on PuzzleVQA and AlgoPuzzleVQA, followed by VisualPuzzles, while VisuLogic remains the most challenging. This consistency suggests that substantial differences in task difficulty persist across models. The supplement reports category-level results.

\subsection{Case study}
Figure~\ref{fig:current-direct-case} illustrates how state construction and execution jointly support puzzle solving. Direct CoT mislocates the goal and overlooks a wall, leading it to propose two downward pushes. Vis2State instead constructs an explicit state containing the player, box, goal, and walls, then searches for a solution under the puzzle's movement constraints. The search checks whether the player can reach the position required for each push and identifies a minimum of four pushes. This example highlights the complementary roles of the two stages. State construction establishes the spatial premises of the problem, while execution checks legal transitions and systematically searches for the shortest solution.

\section{Limitations and Future Work}
\label{sec:limitations}

\textbf{State validity.}
An explicit state may encode incorrect observations or rules, and generated programs may overlook relevant state fields, as discussed in Sec.~\ref{sec:actual-state-use}. Successful execution alone therefore does not guarantee a correct interpretation or solution. Future work should verify states and inferred rules against the image and task constraints, using detected inconsistencies to guide further inspection.

\textbf{Computational efficiency.}
Vis2State uses 1.7--2.8$\times$ the mean output tokens of Inspection + Execution. See the supplement for details. Selective inspection and reuse of previously constructed states may reduce this overhead.

\section{Conclusion}
Vis2State makes explicit problem states the interface between visual understanding and computational reasoning. Its central contribution is to encode the objects, relations, and constraints underlying a solution as a representation that can be inspected, revised, and used by generated programs. Experiments across four benchmarks and two models, supported by component ablations and state interventions, demonstrate the effectiveness of this design and provide evidence that generated programs use the supplied states. These findings highlight explicit state construction as a useful design principle for visual reasoning systems that combine perception with computation.

\label{paper:main-end}

\label{paper:llm-usage-start}
\section*{LLM Usage Statement}
OpenAI Codex and ChatGPT assisted with code development, literature organization, and manuscript revision. OpenAI image-generation tools were used to create the conceptual illustrations. The authors manually checked all content and details in the main paper and supplementary material, including the text, equations, experimental results, figures, tables, and references. The authors take full responsibility for the accuracy and integrity of the work.

\label{paper:references-start}
{\small
\bibliographystyle{ieeenat_fullname}
\bibliography{references}

\begin{thebibliography}{54}
\providecommand{\natexlab}[1]{#1}
\providecommand{\url}[1]{\texttt{#1}}
\expandafter\ifx\csname urlstyle\endcsname\relax
  \providecommand{\doi}[1]{doi: #1}\else
  \providecommand{\doi}{doi: \begingroup \urlstyle{rm}\Url}\fi

\bibitem[{Anthropic}(2024)]{claude3}
{Anthropic}.
\newblock The {Claude 3} model family: {Opus}, {Sonnet}, {Haiku}.
\newblock Technical report, Anthropic, 2024.

\bibitem[{Anthropic}(2025{\natexlab{a}})]{claude37}
{Anthropic}.
\newblock {Claude 3.7 Sonnet System Card}.
\newblock \url{https://www-cdn.anthropic.com/9ff93dfa8f445c932415d335c88852ef47f1201e.pdf}, 2025{\natexlab{a}}.

\bibitem[{Anthropic}(2025{\natexlab{b}})]{claude41}
{Anthropic}.
\newblock {System Card Addendum: Claude Opus 4.1}.
\newblock \url{https://assets.anthropic.com/m/4c024b86c698d3d4/original/Claude-4-1-System-Card.pdf}, 2025{\natexlab{b}}.

\bibitem[{Anthropic}(2026)]{claude47}
{Anthropic}.
\newblock System card: {Claude Opus 4.7}.
\newblock Technical report, Anthropic, 2026.

\bibitem[Bai et~al.(2025{\natexlab{a}})Bai, Cai, Chen, Chen, Chen, Cheng, Deng, Ding, Gao, Ge, et~al.]{qwen3vlreport}
Shuai Bai, Yuxuan Cai, Ruizhe Chen, Keqin Chen, Xionghui Chen, Zesen Cheng, Lianghao Deng, Wei Ding, Chang Gao, Chunjiang Ge, et~al.
\newblock Qwen3-vl technical report.
\newblock \emph{arXiv preprint arXiv:2511.21631}, 2025{\natexlab{a}}.

\bibitem[Bai et~al.(2025{\natexlab{b}})Bai, Chen, Liu, Wang, Ge, Song, Dang, Wang, Wang, Tang, Zhong, Zhu, Yang, Li, Wan, Wang, Ding, Fu, Xu, Ye, Zhang, Xie, Cheng, Zhang, Yang, Xu, and Lin]{qwen25vl}
Shuai Bai, Keqin Chen, Xuejing Liu, Jialin Wang, Wenbin Ge, Sibo Song, Kai Dang, Peng Wang, Shijie Wang, Jun Tang, Humen Zhong, Yuanzhi Zhu, Mingkun Yang, Zhaohai Li, Jianqiang Wan, Pengfei Wang, Wei Ding, Zheren Fu, Yiheng Xu, Jiabo Ye, Xi Zhang, Tianbao Xie, Zesen Cheng, Hang Zhang, Zhibo Yang, Haiyang Xu, and Junyang Lin.
\newblock {Qwen2.5-VL Technical Report}.
\newblock \emph{arXiv preprint arXiv:2502.13923}, 2025{\natexlab{b}}.

\bibitem[{ByteDance Seed}(2026)]{seed21card}
{ByteDance Seed}.
\newblock {Seed2.1} model card: Agentic intelligence for productivity.
\newblock \url{https://lf3-static.bytednsdoc.com/obj/eden-cn/lapzild-tss/ljhwZthlaukjlkulzlp/seed2.1/Seed2_1_Model_Card.pdf}, 2026.

\bibitem[Chen et~al.(2023)Chen, Ma, Wang, and Cohen]{pot}
Wenhu Chen, Xueguang Ma, Xinyi Wang, and William~W Cohen.
\newblock Program of thoughts prompting: Disentangling computation from reasoning for numerical reasoning tasks.
\newblock \emph{Transactions on Machine Learning Research}, 2023.

\bibitem[Chia et~al.(2024)Chia, Toh, Ghosal, Bing, and Poria]{puzzlevqa}
Yew~Ken Chia, Vernon Toh, Deepanway Ghosal, Lidong Bing, and Soujanya Poria.
\newblock {P}uzzle{VQA}: Diagnosing multimodal reasoning challenges of language models with abstract visual patterns.
\newblock In \emph{Findings of the {A}ssociation for {C}omputational {L}inguistics: {ACL} 2024}, pages 16259--16273, 2024.

\bibitem[Comanici et~al.(2025)Comanici, Bieber, Schaekermann, Pasupat, Sachdeva, Dhillon, Blistein, Ram, Zhang, Rosen, et~al.]{gemini25}
Gheorghe Comanici, Eric Bieber, Mike Schaekermann, Ice Pasupat, Noveen Sachdeva, Inderjit Dhillon, Marcel Blistein, Ori Ram, Dan Zhang, Evan Rosen, et~al.
\newblock Gemini 2.5: Pushing the frontier with advanced reasoning, multimodality, long context, and next generation agentic capabilities.
\newblock \emph{arXiv preprint arXiv:2507.06261}, 2025.

\bibitem[Dong et~al.(2026)Dong, Sahin, Wang, Cai, Shrestha, Yang, and Modolo]{toolsrl}
Qihua Dong, Gozde Sahin, Pei Wang, Zhaowei Cai, Robik Shrestha, Hao Yang, and Davide Modolo.
\newblock {Visual Reasoning Through Tool-Supervised Reinforcement Learning}.
\newblock In \emph{Proceedings of the IEEE/CVF Conference on Computer Vision and Pattern Recognition (CVPR) Findings}, pages 8993--9002, 2026.

\bibitem[Fu et~al.(2025)Fu, Liu, Yang, Corring, Lu, Yang, Roth, Florencio, and Zhang]{refocus}
Xingyu Fu, Minqian Liu, Zhengyuan Yang, John~Richard Corring, Yijuan Lu, Jianwei Yang, Dan Roth, Dinei Florencio, and Cha Zhang.
\newblock {ReFocus: Visual Editing as a Chain of Thought for Structured Image Understanding}.
\newblock In \emph{Int. Conf. Mach. Learn.}, pages 17783--17805, 2025.

\bibitem[Gao et~al.(2023)Gao, Madaan, Zhou, Alon, Liu, Yang, Callan, and Neubig]{pal}
Luyu Gao, Aman Madaan, Shuyan Zhou, Uri Alon, Pengfei Liu, Yiming Yang, Jamie Callan, and Graham Neubig.
\newblock Pal: Program-aided language models.
\newblock In \emph{Int. Conf. Mach. Learn.}, pages 10764--10799, 2023.

\bibitem[Ghosal et~al.(2025)Ghosal, Toh, Chia, and Poria]{algopuzzlevqa}
Deepanway Ghosal, Vernon Toh, Yew~Ken Chia, and Soujanya Poria.
\newblock {A}lgo{P}uzzle{VQA}: Diagnosing multimodal reasoning challenges of language models with algorithmic multimodal puzzles.
\newblock In \emph{Proceedings of the 2025 Conference of the Nations of the {A}mericas Chapter of the {A}ssociation for {C}omputational {L}inguistics: Human Language Technologies}, pages 9615--9632, 2025.

\bibitem[{Google DeepMind}(2026{\natexlab{a}})]{gemini31card}
{Google DeepMind}.
\newblock {Gemini 3.1 Pro: Model Card}.
\newblock \url{https://deepmind.google/models/model-cards/gemini-3-1-pro/}, 2026{\natexlab{a}}.

\bibitem[{Google DeepMind}(2026{\natexlab{b}})]{gemini3card}
{Google DeepMind}.
\newblock {Gemini 3 Pro: Model Card}.
\newblock \url{https://storage.googleapis.com/deepmind-media/Model-Cards/Gemini-3-Pro-Model-Card.pdf}, 2026{\natexlab{b}}.

\bibitem[Gupta and Kembhavi(2023)]{visprog}
Tanmay Gupta and Aniruddha Kembhavi.
\newblock Visual programming: Compositional visual reasoning without training.
\newblock In \emph{IEEE Conf. Comput. Vis. Pattern Recog.}, pages 14953--14962. IEEE, 2023.

\bibitem[Hou et~al.(2026)Hou, Xu, Biyani, Li, Liu, Hollon, and Wang]{codev}
Xinhai Hou, Shaoyuan Xu, Manan Biyani, Moyan Li, Jia Liu, Todd~C Hollon, and Bryan Wang.
\newblock Codev: Code with images for faithful visual reasoning via tool-aware policy optimization.
\newblock In \emph{IEEE Conf. Comput. Vis. Pattern Recog.}, pages 21500--21510, 2026.

\bibitem[Hu et~al.(2024)Hu, Shi, Fu, Roth, Ostendorf, Zettlemoyer, Smith, and Krishna]{sketchpad}
Yushi Hu, Weijia Shi, Xingyu Fu, Dan Roth, Mari Ostendorf, Luke Zettlemoyer, Noah~A Smith, and Ranjay Krishna.
\newblock Visual sketchpad: Sketching as a visual chain of thought for multimodal language models.
\newblock \emph{Adv. Neural Inform. Process. Syst.}, pages 139348--139379, 2024.

\bibitem[Hurst et~al.(2024)Hurst, Lerer, Goucher, Perelman, Ramesh, Clark, Ostrow, Welihinda, Hayes, Radford, et~al.]{gpt4ocard}
Aaron Hurst, Adam Lerer, Adam~P Goucher, Adam Perelman, Aditya Ramesh, Aidan Clark, AJ Ostrow, Akila Welihinda, Alan Hayes, Alec Radford, et~al.
\newblock Gpt-4o system card.
\newblock \emph{arXiv preprint arXiv:2410.21276}, 2024.

\bibitem[Johnson et~al.(2017)Johnson, Hariharan, Van Der~Maaten, Hoffman, Fei-Fei, Lawrence~Zitnick, and Girshick]{iep}
Justin Johnson, Bharath Hariharan, Laurens Van Der~Maaten, Judy Hoffman, Li Fei-Fei, C Lawrence~Zitnick, and Ross Girshick.
\newblock Inferring and executing programs for visual reasoning.
\newblock In \emph{Int. Conf. Comput. Vis.}, pages 2989--2998, 2017.

\bibitem[Kavukcuoglu(2025)]{gemini20}
Koray Kavukcuoglu.
\newblock {Gemini 2.0 is Now Available to Everyone}.
\newblock \url{https://blog.google/innovation-and-ai/models-and-research/google-deepmind/gemini-model-updates-february-2025/}, 2025.

\bibitem[Kojima et~al.(2022)Kojima, Gu, Reid, Matsuo, and Iwasawa]{zeroshotcot}
Takeshi Kojima, Shixiang~Shane Gu, Machel Reid, Yutaka Matsuo, and Yusuke Iwasawa.
\newblock {Large Language Models are Zero-Shot Reasoners}.
\newblock \emph{Adv. Neural Inform. Process. Syst.}, 35:\penalty0 22199--22213, 2022.

\bibitem[Li et~al.(2024)Li, Liang, Zeng, Chen, Hausman, Sadigh, Levine, Fei-Fei, Xia, and Ichter]{chaincode}
Chengshu Li, Jacky Liang, Andy Zeng, Xinyun Chen, Karol Hausman, Dorsa Sadigh, Sergey Levine, Li Fei-Fei, Fei Xia, and Brian Ichter.
\newblock Chain of code: Reasoning with a language model-augmented code emulator.
\newblock In \emph{Int. Conf. Mach. Learn.}, pages 28259--28277, 2024.

\bibitem[Lyu et~al.(2023)Lyu, Havaldar, Stein, Zhang, Rao, Wong, Apidianaki, and Callison-Burch]{faithfulcot}
Qing Lyu, Shreya Havaldar, Adam Stein, Li Zhang, Delip Rao, Eric Wong, Marianna Apidianaki, and Chris Callison-Burch.
\newblock {Faithful Chain-of-Thought Reasoning}.
\newblock In \emph{Proceedings of the 13th International Joint Conference on Natural Language Processing and the 3rd Conference of the {A}sia-Pacific Chapter of the {A}ssociation for {C}omputational {L}inguistics}, pages 305--329, 2023.

\bibitem[Mao et~al.(2019)Mao, Gan, Kohli, Tenenbaum, and Wu]{nscl}
Jiayuan Mao, Chuang Gan, Pushmeet Kohli, Joshua~B Tenenbaum, and Jiajun Wu.
\newblock {The Neuro-Symbolic Concept Learner: Interpreting Scenes, Words, and Sentences From Natural Supervision}.
\newblock In \emph{Int. Conf. Learn. Represent.}, 2019.

\bibitem[{OpenAI}(2023)]{gpt4vcard}
{OpenAI}.
\newblock {GPT-4V(ision) System Card}.
\newblock \url{https://openai.com/index/gpt-4v-system-card/}, 2023.

\bibitem[{OpenAI}(2024)]{gpt4omini}
{OpenAI}.
\newblock {GPT-4o mini: Advancing Cost-Efficient Intelligence}.
\newblock \url{https://openai.com/index/gpt-4o-mini-advancing-cost-efficient-intelligence/}, 2024.

\bibitem[{OpenAI}(2025{\natexlab{a}})]{gpt52card}
{OpenAI}.
\newblock {Update to GPT-5 System Card: GPT-5.2}.
\newblock \url{https://openai.com/index/gpt-5-system-card-update-gpt-5-2/}, 2025{\natexlab{a}}.

\bibitem[{OpenAI}(2025{\natexlab{b}})]{o3o4card}
{OpenAI}.
\newblock {OpenAI o3 and o4-mini System Card}.
\newblock Technical report, OpenAI, 2025{\natexlab{b}}.

\bibitem[{OpenAI}(2026)]{gpt55card}
{OpenAI}.
\newblock {GPT-5.5 System Card}.
\newblock \url{https://openai.com/index/gpt-5-5-system-card/}, 2026.

\bibitem[Singh et~al.(2025)Singh, Fry, Perelman, Tart, Ganesh, El-Kishky, McLaughlin, Low, Ostrow, Ananthram, et~al.]{gpt5card}
Aaditya Singh, Adam Fry, Adam Perelman, Adam Tart, Adi Ganesh, Ahmed El-Kishky, Aidan McLaughlin, Aiden Low, AJ Ostrow, Akhila Ananthram, et~al.
\newblock Openai gpt-5 system card.
\newblock \emph{arXiv preprint arXiv:2601.03267}, 2025.

\bibitem[Song et~al.(2026{\natexlab{a}})Song, Li, Yu, Zhou, Yang, Bai, She, Huang, and Zhao]{codedance}
Qi Song, Honglin Li, Yingchen Yu, Haoyi Zhou, Lin Yang, Song Bai, Qi She, Zilong Huang, and Yunqing Zhao.
\newblock Codedance: A dynamic tool-integrated mllm for executable visual reasoning.
\newblock In \emph{IEEE Conf. Comput. Vis. Pattern Recog.}, pages 19186--19195, 2026{\natexlab{a}}.

\bibitem[Song et~al.(2026{\natexlab{b}})Song, Ou, Kong, Li, Neubig, and Yue]{visualpuzzles}
Yueqi Song, Tianyue Ou, Yibo Kong, Zecheng Li, Graham Neubig, and Xiang Yue.
\newblock {VisualPuzzles: Decoupling Multimodal Reasoning Evaluation from Domain Knowledge}.
\newblock In \emph{Int. Conf. Mach. Learn.}, pages 114860--114932, 2026{\natexlab{b}}.

\bibitem[Su et~al.(2026)Su, Wang, Ren, Lin, and Chen]{pixelreasoner}
Alex Su, Haozhe Wang, Weiming Ren, Fangzhen Lin, and Wenhu Chen.
\newblock Pixel reasoner: Incentivizing pixel-space reasoning with curiosity-driven reinforcement learning.
\newblock \emph{Adv. Neural Inform. Process. Syst.}, 38:\penalty0 8222--8251, 2026.

\bibitem[Sur{\'\i}s et~al.(2023)Sur{\'\i}s, Menon, and Vondrick]{viper}
D{\'\i}dac Sur{\'\i}s, Sachit Menon, and Carl Vondrick.
\newblock Vipergpt: Visual inference via python execution for reasoning.
\newblock In \emph{Int. Conf. Comput. Vis.}, pages 11854--11864. IEEE, 2023.

\bibitem[Toh et~al.(2026)Toh, Chia, Ghosal, and Poria]{tracking2026}
Vernon Toh, Yew~Ken Chia, Deepanway Ghosal, and Soujanya Poria.
\newblock Tracking the evolution of multimodal reasoning on visual puzzles.
\newblock In \emph{Logical and Symbolic Reasoning in Language Models, AAAI 2026}, 2026.

\bibitem[{Volcano Engine}(2025)]{doubao16vision}
{Volcano Engine}.
\newblock {Doubao-Seed-1.6-vision}.
\newblock \url{https://docs.volcengine.com/docs/ark/model-release-announcement-2025?lang=en}, 2025.

\bibitem[Wang et~al.(2024{\natexlab{a}})Wang, Bai, Tan, Wang, Fan, Bai, Chen, Liu, Wang, Ge, et~al.]{qwen2vl}
Peng Wang, Shuai Bai, Sinan Tan, Shijie Wang, Zhihao Fan, Jinze Bai, Keqin Chen, Xuejing Liu, Jialin Wang, Wenbin Ge, et~al.
\newblock Qwen2-vl: Enhancing vision-language model's perception of the world at any resolution.
\newblock \emph{arXiv preprint arXiv:2409.12191}, 2024{\natexlab{a}}.

\bibitem[Wang et~al.(2026)Wang, Shi, Cheng, Zhang, He, Tang, Zhang, Yu, Chen, Xu, et~al.]{beacon}
Qixun Wang, Yang Shi, Letian Cheng, Zhuoran Zhang, Yan He, Yuqi Tang, Qi Zhang, Xinlei Yu, Ruizhe Chen, Tianrun Xu, et~al.
\newblock Beacon: Knowing when and how to perform agentic visual reasoning.
\newblock \emph{arXiv preprint arXiv:2607.28595}, 2026.

\bibitem[Wang et~al.(2024{\natexlab{b}})Wang, Chen, Yuan, Zhang, Li, Peng, and Ji]{codeact}
Xingyao Wang, Yangyi Chen, Lifan Yuan, Yizhe Zhang, Yunzhu Li, Hao Peng, and Heng Ji.
\newblock {Executable Code Actions Elicit Better LLM Agents}.
\newblock In \emph{Int. Conf. Mach. Learn.}, pages 50208--50232, 2024{\natexlab{b}}.

\bibitem[Wei et~al.(2022)Wei, Wang, Schuurmans, Bosma, Xia, Chi, Le, Zhou, et~al.]{cot}
Jason Wei, Xuezhi Wang, Dale Schuurmans, Maarten Bosma, Fei Xia, Ed Chi, Quoc~V Le, Denny Zhou, et~al.
\newblock Chain-of-thought prompting elicits reasoning in large language models.
\newblock \emph{Adv. Neural Inform. Process. Syst.}, pages 24824--24837, 2022.

\bibitem[Wu et~al.(2026{\natexlab{a}})Wu, Yang, Jiang, Li, Yan, Yu, Zhang, Zhai, and Nahrstedt]{vtoolr1}
Mingyuan Wu, Jingcheng Yang, Jize Jiang, Meitang Li, Kaizhuo Yan, Hanchao Yu, Minjia Zhang, Chengxiang Zhai, and Klara Nahrstedt.
\newblock Vtool-r1: Vlms learn to think with images via reinforcement learning on multimodal tool use.
\newblock In \emph{Int. Conf. Learn. Represent.}, pages 78298--78319, 2026{\natexlab{a}}.

\bibitem[Wu and Xie(2024)]{vstar}
Penghao Wu and Saining Xie.
\newblock V*: Guided visual search as a core mechanism in multimodal llms.
\newblock In \emph{IEEE Conf. Comput. Vis. Pattern Recog.}, pages 13084--13094, 2024.

\bibitem[Wu et~al.(2026{\natexlab{b}})Wu, Hua, Deng, and Ye]{vctooler}
Yizheng Wu, Jiashen Hua, Bing Deng, and Jieping Ye.
\newblock Vc-tooler: Learning compositional and adaptive visual tool use.
\newblock \emph{arXiv preprint arXiv:2608.02217}, 2026{\natexlab{b}}.

\bibitem[Xu et~al.(2026{\natexlab{a}})Xu, Li, Lin, Xue, Xian, Xu, Wu, Zhang, Deng, Yu, et~al.]{deepseek41}
Anyi Xu, B Li, Bangcai Lin, Bing Xue, BingCheng Xian, Bingzheng Xu, Bochao Wu, Bowei Zhang, Boyi Deng, CC Yu, et~al.
\newblock {DeepSeek-V4.1-Flash}: Pushing the limits of {KV} cache compression.
\newblock \emph{arXiv preprint arXiv:2609.19969}, 2026{\natexlab{a}}.

\bibitem[Xu et~al.(2026{\natexlab{b}})Xu, Wang, Wang, Chen, Zhou, Yang, Lu, Li, Wang, Zhu, et~al.]{visulogic}
Weiye Xu, Jiahao Wang, Weiyun Wang, Zhe Chen, Wengang Zhou, Aijun Yang, Lewei Lu, Houqiang Li, Xiaohua Wang, Xizhou Zhu, et~al.
\newblock Visulogic: A benchmark for evaluating visual reasoning in multi-modal large language models.
\newblock In \emph{Int. Conf. Learn. Represent.}, pages 25966--26003, 2026{\natexlab{b}}.

\bibitem[Yan et~al.(2026)Yan, Yan, Xu, Zhou, Chen, Ji, Cao, Zhang, Liu, Zhang, et~al.]{seetothink}
Siyu Yan, Zhuoran Yan, Haiying Xu, Panhao Zhou, Jingyu Chen, Chenhao Ji, Shuo Cao, Yongheng Zhang, Haoze Liu, Siyu Zhang, et~al.
\newblock See2think: Do multimodal models really use intermediate visual states?
\newblock \emph{arXiv preprint arXiv:2607.26769}, 2026.

\bibitem[Yao et~al.(2023)Yao, Zhao, Yu, Du, Shafran, Narasimhan, and Cao]{react}
Shunyu Yao, Jeffrey Zhao, Dian Yu, Nan Du, Izhak Shafran, Karthik Narasimhan, and Yuan Cao.
\newblock React: Synergizing reasoning and acting in language models.
\newblock In \emph{Int. Conf. Learn. Represent.}, 2023.

\bibitem[Yi et~al.(2018)Yi, Wu, Gan, Torralba, Kohli, and Tenenbaum]{nsvqa}
Kexin Yi, Jiajun Wu, Chuang Gan, Antonio Torralba, Pushmeet Kohli, and Josh Tenenbaum.
\newblock Neural-symbolic vqa: Disentangling reasoning from vision and language understanding.
\newblock In \emph{Adv. Neural Inform. Process. Syst.}, 2018.

\bibitem[Zeng et~al.(2026)Zeng, Lv, Hou, Du, Zheng, Chen, Yin, Ge, Huang, Xie, et~al.]{glm5}
Aohan Zeng, Xin Lv, Zhenyu Hou, Zhengxiao Du, Qinkai Zheng, Bin Chen, Da Yin, Chendi Ge, Chenghua Huang, Chengxing Xie, et~al.
\newblock {GLM-5: from Vibe Coding to Agentic Engineering}.
\newblock \emph{arXiv preprint arXiv:2602.15763}, 2026.

\bibitem[Zhang et~al.(2026)Zhang, Lu, Yin, Fu, Chen, Hu, Wen, Jiang, Liu, Zhang, Fan, Chen, Chen, Ding, Tang, Zhang, Wang, Yang, Gao, and Zhou]{thyme}
Yi-Fan Zhang, Xingyu Lu, Shukang Yin, Chaoyou Fu, Wei Chen, Xiao Hu, Bin Wen, Kaiyu Jiang, Changyi Liu, Tianke Zhang, Haonan Fan, Kaibing Chen, Jiankang Chen, Haojie Ding, Kaiyu Tang, Zhang Zhang, Liang Wang, Fan Yang, Tingting Gao, and Guorui Zhou.
\newblock {Thyme: Think Beyond Images}.
\newblock In \emph{Int. Conf. Learn. Represent.}, pages 79536--79572, 2026.

\bibitem[Zheng et~al.(2026)Zheng, Yang, Hong, Zhao, Xu, Yang, and Shen]{deepeyes}
Ziwei Zheng, Minghao Yang, Jack Hong, Chenxiao Zhao, Guohai Xu, Le Yang, and Chao Shen.
\newblock Deepeyes: Incentivizing "thinking with images" via reinforcement learning.
\newblock In \emph{Int. Conf. Learn. Represent.}, pages 126775--126798, 2026.

\bibitem[Zhu et~al.(2025)Zhu, Wang, Chen, Liu, Ye, Gu, Tian, Duan, Su, Shao, et~al.]{internvl3}
Jinguo Zhu, Weiyun Wang, Zhe Chen, Zhaoyang Liu, Shenglong Ye, Lixin Gu, Hao Tian, Yuchen Duan, Weijie Su, Jie Shao, et~al.
\newblock Internvl3: Exploring advanced training and test-time recipes for open-source multimodal models.
\newblock \emph{arXiv preprint arXiv:2504.10479}, 2025.

\end{thebibliography}
}

\clearpage

\appendix
\setcounter{table}{0}
\renewcommand{\thetable}{S\arabic{table}}
\setcounter{figure}{0}
\renewcommand{\thefigure}{S\arabic{figure}}
\setcounter{section}{0}
\renewcommand{\thesection}{S\arabic{section}}
\setcounter{algorithm}{0}
\renewcommand{\thealgorithm}{S\arabic{algorithm}}

\renewcommand{\theHsection}{supp.\arabic{section}}
\renewcommand{\theHsubsection}{\theHsection.\arabic{subsection}}
\renewcommand{\theHsubsubsection}{\theHsubsection.\arabic{subsubsection}}
\renewcommand{\theHfigure}{supp.\arabic{figure}}
\renewcommand{\theHtable}{supp.\arabic{table}}
\def\theHalgorithm{supp.\arabic{algorithm}}

\preprintsupplement

\section{Algorithm and runtime}
\label{sec:additional-protocol}

\begin{algorithm}[htbp]
\caption{Core reasoning loop of Vis2State}
\label{alg:vis2state}
\footnotesize
\begin{algorithmic}[1]
\Require Image $x$, question $q$, budgets $B,T$
\State $H_0\gets[\,]$
\For{$t=1,\ldots,B$ while elapsed time $<T$}
  \State $a_t\gets M_\theta(x,q,H_{t-1};\mathcal{P})$ \Comment{Choose next action}
  \If{$a_t$ submits a final answer $y$}
    \State \Return $y$
  \Else
    \State Perform $a_t$ and receive $o_t$ \Comment{Inspect or execute}
  \EndIf
  \State $H_t\gets H_{t-1}\mathbin{\|}[(a_t,o_t)]$ \Comment{Append feedback}
\EndFor
\State \Return no submission
\end{algorithmic}
\end{algorithm}

Algorithm~\ref{alg:vis2state} summarizes the Vis2State reasoning loop. Given an image $x$ and question $q$, a vision-language model $M_\theta$ follows the shared prompt $\mathcal{P}$ and selects each action $a_t$ using the history $H_{t-1}$ of prior actions and observations. The history starts empty. Inspection gathers visual evidence for constructing or refining the state, while execution runs a program on the supplied state. Each action and its returned observation $o_t$ are appended to the history, allowing the model to revise the state or program before choosing its next action. The loop ends when the model submits a final answer or exhausts either budget.

Each attempt allows at most $B=22$ model calls within $T=1{,}800\,\mathrm{s}$ of active runtime. Table~\ref{tab:runtime-settings} summarizes these budgets and the limits on individual requests and executions. Each Python execution is limited to 3 seconds and 256 MB of memory. The program receives the supplied JSON state without access to the source image, while visual observations remain available to the model.

\begin{table}[htbp]
\centering\small
\setlength{\tabcolsep}{3pt}
\caption{\textbf{Runtime settings.} Request timeouts are also bounded by the remaining per-attempt time budget.}
\label{tab:runtime-settings}
\begin{tabular*}{\linewidth}{@{\extracolsep{\fill}}lr@{}}
\toprule
Setting & Limit \\
\midrule
Model calls per attempt ($B$) & 22 \\
Active runtime per attempt ($T$) & 1,800 s \\
Output tokens per model call & 65,536 \\
HTTP timeout per request & 1,200 s \\
Python time per execution & 3 s \\
Python memory per execution & 256 MB \\
\bottomrule
\end{tabular*}
\end{table}

\paragraph{Step-error feedback.}
Tool errors are returned as observations for subsequent reasoning. Within a Vis2State attempt, a length-truncated response consumes a step and returns error feedback without executing partial content. Prior observations and execution results remain in the history, and the model can issue another action within the remaining call and time budgets. A final answer requires explicit submission by the model.

\section{Vis2State Prompt}
\label{sec:shared-prompt}
Table~\ref{tab:shared-prompt} summarizes the shared prompt and execution interface used across all datasets and models. It retains the state-construction, verification, execution, and submission requirements while omitting the full action schema and tool argument specifications.

\begin{table*}[t]
\centering
\caption{\textbf{Vis2State prompt and execution interface.}}
\label{tab:shared-prompt}
\setlength{\fboxsep}{7pt}
\fbox{\begin{minipage}{\dimexpr\textwidth-2\fboxsep-2\fboxrule\relax}
\small
{\large\bfseries Vis2State prompt\par}\smallskip
Solve the puzzle from the image and question by constructing, checking, and solving an explicit state.

\textbf{State construction.} Choose a minimal JSON object containing the facts, relations, rules, and objective needed to answer. Define its fields and structure yourself without a predefined puzzle schema. Distinguish observed facts from assumptions, and preserve relevant coordinate conventions and action preconditions.

\textbf{Visual inspection.} Use source-image tools to check answer-relevant facts in the proposed state. Compare their actual outputs and the original image with the representation, and repair missing or inconsistent facts. A successful measurement does not establish semantic correctness. Resolve uncertainty that could change the answer before solving.

\textbf{Execution and revision.} Write Python defining \texttt{solve(state)} to read the checked state and return a JSON-compatible result. Put the JSON-encoded input object in \texttt{state\_json} and the program in \texttt{source}. The program receives this object unchanged, without image access. Copy observed values into the state rather than referring to measurement outputs. Verify the task rules and objective in the computation. If the state changes, recompute affected results.

\textbf{Action format.} Return one JSON action matching the supplied schema. Use \texttt{measure} to inspect, \texttt{code} to compute, \texttt{final} to submit, or \texttt{unsupported} if essential evidence remains unresolved. Supply \texttt{state\_json} only with \texttt{code}, and set it to an empty string for other actions. Set unused text fields to empty strings and unused \texttt{requests} to \texttt{[]}. Briefly explain each action in \texttt{reason}.

\textbf{Answer submission.} Inspect tool feedback before submitting. Map the computed result to the requested answer. For the multiple-choice benchmarks, \texttt{answer\_json} must encode one displayed uppercase option letter.
\end{minipage}}
\end{table*}

\section{Evaluation protocol}
\label{sec:appendix-evaluation}

\subsection{Answer extraction and scoring}
We use each benchmark's released answer-extraction and scoring procedures~\cite{algopuzzlevqa,visualpuzzles,visulogic,puzzlevqa}. Direct CoT follows the official CoT prompt for each benchmark and answers without tools. Its responses undergo answer extraction, whereas tool-assisted methods submit an option through the final-answer interface. For VisuLogic, DeepSeek V4.1 Flash replaces the released evaluator's GPT-4o-mini because of model availability. PuzzleVQA averages its four single-concept and six dual-concept categories, weighting the group averages by 0.4 and 0.6, respectively.

\paragraph{Missing answers and attempt selection.}
Uniform random completion applies only to final records marked as having no prediction. Existing complete predictions are preserved, including incorrect or invalid answers. For each method--dataset pair, we reset the random seed to 0, order questions by ID, and sample uniformly from each question's available answer options without consulting the reference answer. Valid incorrect answers are not retried.

\subsection{Statistical comparisons}
We compare Vis2State and Direct CoT on the same questions within each benchmark. For each question, we encode the difference in correctness as $+1$, $-1$, or $0$. We compute the full empirical paired bootstrap distribution of the mean difference by convolving these three outcome probabilities over the dataset size. The 2.5th and 97.5th percentiles give the reported 95.0\% confidence interval.

We test the paired disagreements with a two-sided exact McNemar test and apply Holm correction across the four benchmarks. These estimates reflect question-level uncertainty within a single run and do not account for correlations among questions sharing an image.

\subsection{Sources of published comparisons}
\label{sec:recent-reference-protocols}
The sources below provide published scores for each benchmark. Their model versions and evaluation protocols differ from our experiments, so these results provide context for the comparisons between Vis2State and Direct CoT using the same model.

\paragraph{AlgoPuzzleVQA results.}
GPT-4V, Claude 3 Opus, and Qwen2-VL-7B are from~\cite{algopuzzlevqa}. Gemini 3 Pro Preview, o4-mini (high), o3 (high), GPT-5.2, and Qwen3-VL-8B-Instruct are from~\cite{tracking2026}.

\paragraph{VisualPuzzles results.}
Human (50th Percentile), o4-mini, o3, Gemini 2.5 Pro, Claude 3.7 Sonnet, and Qwen2.5-VL-72B-Instruct are from~\cite{visualpuzzles}. GPT-5 (high), Qwen3-VL-235B-A22B (thinking/instruct), and Claude Opus 4.1 (thinking) are from~\cite{qwen3vlreport}. Gemini 3.1 Pro and Qwen3-VL-8B-Instruct are from~\cite{beacon}.

\paragraph{VisuLogic results.}
Human, o3 (20250417), GPT-4o (20240806), Doubao 1.6 Vision (250815), Gemini 2.0 Pro (20250205), Qwen2.5-VL-72B, and InternVL3-78B are from~\cite{visulogic}. Gemini 3.1 Pro, GPT-5.5, and Claude Opus 4.7 are from~\cite{seed21card}. Additional thinking/instruct/high-mode results are from~\cite{qwen3vlreport}.

\paragraph{PuzzleVQA results.}
CoT results for GPT-4V and Claude 3 Opus are from~\cite{puzzlevqa}. All other published results are from~\cite{tracking2026}.

\begin{table*}[t]
\centering\footnotesize
\setlength{\tabcolsep}{2pt}
\caption{\textbf{GLM-5.3-Flash on AlgoPuzzleVQA}. Accuracy (\%) across 18 task families. Bold marks the better result, including ties.}
\label{tab:glm-detail-ap}
\begin{tabularx}{\textwidth}{@{}l*{10}{>{\raggedleft\arraybackslash}X}@{}}
\toprule
Method & Overall & Board & Calendar & Chain & Checker & Clock & Colour & Map & Maze & Move box \\
\midrule
Questions & 1,800 & 100 & 100 & 100 & 100 & 100 & 100 & 100 & 100 & 100 \\
\midrule
Direct CoT & 82.3 & \textbf{100.0} & 99.0 & 56.0 & 92.0 & 86.0 & 40.0 & 65.0 & 56.0 & 84.0 \\
\rowcolor{ComparisonRow}
Vis2State (Ours) & \textbf{94.5} & \textbf{100.0} & \textbf{100.0} & \textbf{74.0} & \textbf{100.0} & \textbf{100.0} & \textbf{99.0} & \textbf{71.0} & \textbf{92.0} & \textbf{98.0} \\
Gain (pp) & +12.2 & 0.0 & +1.0 & +18.0 & +8.0 & +14.0 & +59.0 & +6.0 & +36.0 & +14.0 \\
\bottomrule
\end{tabularx}
\par\vspace{4pt}
\begin{tabularx}{\textwidth}{@{}l*{10}{>{\raggedleft\arraybackslash}X}@{}}
\toprule
Method & Overall & Queens & Number & Fruits & Rubik & Think dot & Hanoi & Water & Wheel & Wood \\
\midrule
Questions & 1,800 & 100 & 100 & 100 & 100 & 100 & 100 & 100 & 100 & 100 \\
\midrule
Direct CoT & 82.3 & \textbf{100.0} & 96.0 & 99.0 & 95.0 & 70.0 & 70.0 & 97.0 & 85.0 & 92.0 \\
\rowcolor{ComparisonRow}
Vis2State (Ours) & \textbf{94.5} & \textbf{100.0} & \textbf{100.0} & \textbf{100.0} & \textbf{99.0} & \textbf{73.0} & \textbf{100.0} & \textbf{99.0} & \textbf{96.0} & \textbf{100.0} \\
Gain (pp) & +12.2 & 0.0 & +4.0 & +1.0 & +4.0 & +3.0 & +30.0 & +2.0 & +11.0 & +8.0 \\
\bottomrule
\end{tabularx}
\end{table*}

\begin{table*}[t]
\centering\footnotesize
\setlength{\tabcolsep}{2pt}
\caption{\textbf{GLM-5.3-Flash on VisualPuzzles}. Accuracy (\%) across five reasoning categories. Bold marks the better result, including ties.}
\label{tab:glm-detail-vp}
\begin{tabularx}{\textwidth}{@{}l*{6}{>{\raggedleft\arraybackslash}X}@{}}
\toprule
Method & Overall & Algorithmic & Analogical & Deductive & Inductive & Spatial \\
\midrule
Questions & 1,168 & 262 & 211 & 200 & 209 & 286 \\
\midrule
Direct CoT & 67.2 & 80.9 & \textbf{81.0} & 83.5 & 35.4 & 56.3 \\
\rowcolor{ComparisonRow}
Vis2State (Ours) & \textbf{75.1} & \textbf{91.2} & \textbf{81.0} & \textbf{86.0} & \textbf{53.1} & \textbf{64.3} \\
Gain (pp) & +7.9 & +10.3 & 0.0 & +2.5 & +17.7 & +8.0 \\
\bottomrule
\end{tabularx}
\end{table*}

\begin{table*}[t]
\centering\footnotesize
\setlength{\tabcolsep}{2pt}
\caption{\textbf{GLM-5.3-Flash on VisuLogic}. Accuracy (\%) across six reasoning categories. Bold marks the better result, including ties.}
\label{tab:glm-detail-vl}
\begin{tabularx}{\textwidth}{@{}l*{7}{>{\raggedleft\arraybackslash}X}@{}}
\toprule
Method & Overall & Quantity & Spatiality & Position & Attribute & Style & Other \\
\midrule
Questions & 1,000 & 353 & 231 & 136 & 82 & 90 & 108 \\
\midrule
Direct CoT & 36.2 & 37.7 & \textbf{29.9} & 45.6 & \textbf{39.0} & 24.4 & 40.7 \\
\rowcolor{ComparisonRow}
Vis2State (Ours) & \textbf{44.2} & \textbf{47.0} & \textbf{29.9} & \textbf{55.9} & 37.8 & \textbf{51.1} & \textbf{50.0} \\
Gain (pp) & +8.0 & +9.3 & 0.0 & +10.3 & -1.2 & +26.7 & +9.3 \\
\bottomrule
\end{tabularx}
\end{table*}

\begin{table*}[t]
\centering\footnotesize
\setlength{\tabcolsep}{2pt}
\caption{\textbf{GLM-5.3-Flash on PuzzleVQA}. Accuracy (\%) across single- and dual-concept categories. Bold marks the better result, including ties. N/C/S/Z denote numbers/colors/shapes/size.}
\label{tab:glm-detail-pq}
\begin{tabularx}{\textwidth}{@{}l*{13}{>{\raggedleft\arraybackslash}X}@{}}
\toprule
 & & \multicolumn{5}{c}{Single-concept} & \multicolumn{7}{c}{Dual-concept} \\
\cmidrule(lr){3-7}\cmidrule(lr){8-14}
Method & Overall & N & C & Z & S & Avg. & N$+$S & C$+$N & N$+$Z & C$+$S & Z$+$S & C$+$Z & Avg. \\
\midrule
Questions & 2,000 & 200 & 200 & 200 & 200 & 800 & 200 & 200 & 200 & 200 & 200 & 200 & 1,200 \\
\midrule
Direct CoT & 89.4 & 93.0 & \textbf{100.0} & 95.0 & 68.5 & 89.1 & 91.0 & 99.0 & 85.0 & 86.0 & 98.5 & 78.0 & 89.6 \\
\rowcolor{ComparisonRow}
Vis2State (Ours) & \textbf{96.3} & \textbf{99.0} & \textbf{100.0} & \textbf{100.0} & \textbf{74.5} & \textbf{93.4} & \textbf{99.5} & \textbf{99.5} & \textbf{94.5} & \textbf{97.5} & \textbf{100.0} & \textbf{98.5} & \textbf{98.3} \\
Gain (pp) & +6.9 & +6.0 & 0.0 & +5.0 & +6.0 & +4.3 & +8.5 & +0.5 & +9.5 & +11.5 & +1.5 & +20.5 & +8.7 \\
\bottomrule
\end{tabularx}
\end{table*}

\section{Detailed cross-model results}
\label{sec:glm-details}
Tables~\ref{tab:glm-detail-ap}--\ref{tab:glm-detail-pq} compare Direct CoT and Vis2State using GLM-5.3-Flash-260828~\cite{glm5} across all four benchmarks. Both methods use the evaluation protocol in Sec.~\ref{sec:appendix-evaluation}. Vis2State uses the same prompt, tool instructions, and state interface as the DeepSeek experiments, with the shared prompt specified in Sec.~\ref{sec:shared-prompt}. The tables report question counts, accuracy, and gains computed before rounding. PuzzleVQA also reports single- and dual-concept averages.

Vis2State improves 33 of 39 categories, ties five, and declines in one. The largest gains within the four benchmarks occur in Colour (+59.0 percentage points), Inductive reasoning (+17.7), Style (+26.7), and colors--size (+20.5), respectively. Benefits are not uniform. VisuLogic Spatiality remains at 29.9\%, while Attribute decreases by 1.2 percentage points. PuzzleVQA Shapes improves to 74.5\%, the lowest of its concept categories.

\section{Supplementary analyses}
\label{sec:appendix-analyses}

\subsection{Accuracy and output-token cost}
\label{sec:token-measurement}
We compare five configurations of DeepSeek V4.1 Flash. \textbf{Direct CoT} answers without tools, \textbf{Inspection only} provides visual tools, and \textbf{Execution only} provides Python execution and feedback. \textbf{Inspection + Execution} combines both tool types without requiring an explicit state. \textbf{Vis2State} adds state-construction guidance and the \texttt{solve(state)} interface.

\begin{figure*}[t]
\centering
\includegraphics[width=0.85\textwidth]{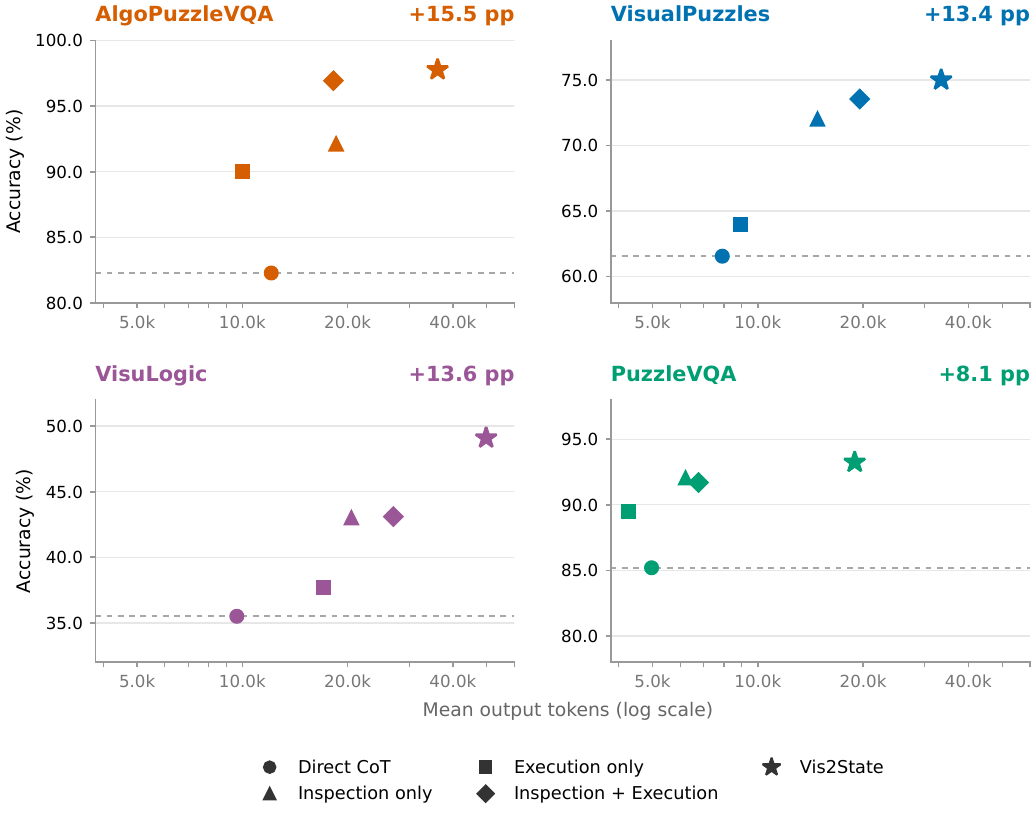}
\caption{\textbf{Accuracy and output-token cost with DeepSeek V4.1 Flash.} Full-benchmark accuracy is plotted against mean output tokens on sampled questions. Panels use a shared logarithmic cost scale and accuracy ranges spanning 20 percentage points. Stars denote Vis2State, dashed lines indicate Direct CoT accuracy, and annotations show the corresponding accuracy gains.}
\label{fig:cost-accuracy}
\end{figure*}

For each dataset, we first identify questions with valid correct answers and complete usage records for all five methods. We then randomly sample 100 questions without replacement from this intersection (seed 0), using the same question IDs for all five methods. Figure~\ref{fig:cost-accuracy} plots mean output tokens over these same IDs against full-benchmark accuracy. Output tokens include reasoning and all solving requests within each selected final attempt, including inspection, code, and step-error feedback. The cost measure is conditional on common correct answers, whereas accuracy is evaluated on the full benchmark.

From Figure~\ref{fig:cost-accuracy}, we observe that Vis2State achieves the highest full-benchmark accuracy in all four panels, while using 1.7--2.8$\times$ the mean output tokens of Inspection + Execution on these matched subsets. The accuracy gains therefore accompany additional output-token cost.

\subsection{Error cohort and annotation}
\label{sec:error-annotation}

\paragraph{Diagnostic cohort.}
We analyze all 1,709 Direct CoT responses scored incorrect by the official evaluators: 319 from AlgoPuzzleVQA, 449 from VisualPuzzles, 645 from VisuLogic, and 296 from PuzzleVQA. Vis2State answers 971 of these questions correctly through valid submissions, excluding random completions. The corresponding correction rates are 92.5\%, 52.1\%, 36.3\%, and 70.3\%, respectively.

\paragraph{Annotation and adjudication.}
We annotate perceptual and reasoning errors using an LLM-as-judge protocol with human adjudication. DeepSeek V4.1 Flash is the primary judge. Cases it cannot clearly classify are referred to GPT-6 Astra. The authors adjudicate cases that remain ambiguous. Judges receive the image, question, full model response, and reference option. Labels cover perception, representation, rule/inference, search/state updates, calculation, and answer comparison. Each label requires a concrete discrepancy supported by a response quotation. For reporting, \textbf{Perception} covers visual errors, \textbf{State / Reasoning} combines representation and rule/inference errors, and \textbf{Search / Computation} combines search/state updates, calculation, and answer comparison. \textbf{Compound} includes cases spanning multiple groups, answer-extraction failures, and otherwise unattributed cases.

\subsection{State intervention protocol}
\label{sec:state-counterfactual-protocol}
We test whether saved programs use the supplied state by changing the state while holding the program fixed.

\paragraph{Subset selection and composition.}
We select 200 questions each from AlgoPuzzleVQA and PuzzleVQA for which the last saved DeepSeek Vis2State program correctly solves the original state and independent reference programs can certify the intervention answers. Original-state accuracy is 100.0\% by construction. Selection does not require the saved program to solve the counterfactual or control correctly.

AlgoPuzzleVQA includes water jugs (61), color rearrangement (39), Tower of Hanoi (31), checker moves (26), rotting kiwi (22), N-queens (17), and board tiling (4). PuzzleVQA includes number grids (55), color grids (32), size grids (30), Venn diagrams (28), color hexagons (21), number triangles (17), polygon side counts (12), and number--color associations (5).

\paragraph{Interventions and independent verification.}
Each question has three state conditions: the original state, an answer-changing counterfactual, and an answer-preserving control. Both interventions modify task-relevant values, positions, or configurations; repeated representations of a changed attribute are updated consistently. Independent reference programs verify that the original answer matches the benchmark reference, the counterfactual is valid and maps to a different existing answer option, and the control preserves the original answer. These checks establish the target answers independently of the saved program under evaluation.

\paragraph{Replay and scoring.}
We replace only the supplied state, keeping the program, answer options, and execution limits fixed, with no new model inference. The 400 questions yield 1,200 condition records. Each output is scored against the independently verified answer for its state condition; when both a value and an option are returned, both must be correct. All intervention failures remain in the denominator, and no question is replaced because of an incorrect intervention output.

\paragraph{Results.}
Counterfactual accuracy is 98.0\% (196/200) on AlgoPuzzleVQA and 93.5\% (187/200) on PuzzleVQA. Control accuracy is 100.0\% (200/200) and 98.5\% (197/200), respectively. Of the 17 counterfactual failures, 9 compute the correct new value but return an inconsistent or missing option; the remaining 8 return incorrect or missing answers. Independent verification certifies the intervention answers, but does not guarantee that the saved program produces them. On these original-correct subsets, the results show that programs usually respond correctly to answer-changing state edits while preserving correctness under answer-preserving edits.

\section{Qualitative examples}
\label{sec:appendix-examples}
We first examine how inspection and execution contribute to recorded solutions, then present task-dependent states and the computations they support. The model generates these states while solving each question, determining their fields, structure, and values without a predefined puzzle-specific schema.

\begin{figure*}[tb]
\centering
\includegraphics[width=0.95\textwidth]{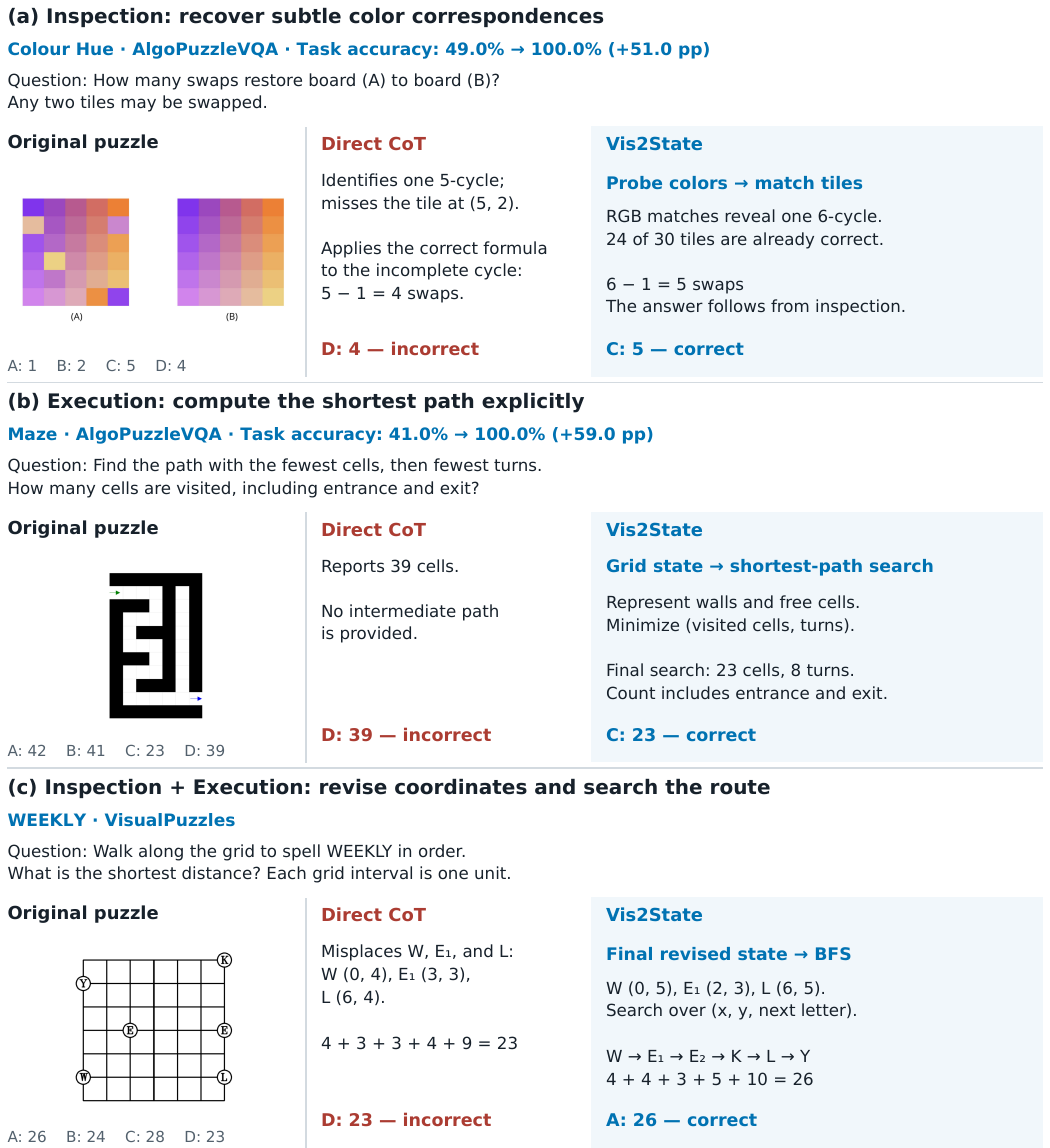}
\caption{\textbf{Inspection, execution, and their combination.} (a) Color probes refine tile correspondences into a six-tile cycle. (b) Search over the maze state computes the shortest path. (c) Revised coordinates and search yield the shortest WEEKLY walk. Task accuracies in (a,b) compare Direct CoT with Vis2State over 100 questions each. Only decisive steps are shown. Coordinates are one-based (row, column) in (a) and zero-based $(x,y)$ from the top-left in (c).}
\label{fig:case-components}
\end{figure*}

\subsection{Inspection and execution in recorded solutions}
\label{sec:current-case-protocol}
Figure~\ref{fig:case-components} illustrates the complementary roles of inspection and execution through three recorded solutions. In Colour Hue, color probes recover tile correspondences and reveal a six-tile cycle, allowing the model to infer five swaps without code execution. In Maze, a program searches the grid state for the path with the fewest visited cells, using the number of turns to break ties. In WEEKLY, the model revises the letter coordinates and searches the resulting state for the shortest walk spelling the word. Together, these examples show how inspection helps establish an accurate problem state, while execution supports systematic search over that state.

\subsection{Task-dependent state representations}
\label{sec:state-examples}
The following six excerpts show model-generated states from recorded trajectories across all four benchmarks. They cover configurations and legal moves, spatial occupancy, discriminative attributes, and measured quantities. Each pairs a state excerpt with the computation it supports. 

\paragraph{Water-jug quantities and target configurations.}
In AlgoPuzzleVQA, each list position refers to the same jug. For example, the first jug has capacity 11, initially contains 1 unit, and must finish with 5 units. All entries use the puzzle's volume unit.
{\small
\begin{verbatim}
{"capacities": [11,9,8,4,3],
 "initial":    [1,1,8,3,1],
 "target":     [5,0,3,3,3]}
\end{verbatim}
}
The program uses capacities to determine legal pours and searches from the initial to the target configuration. The recorded solution takes four moves.

\paragraph{Maze connectivity.}
In AlgoPuzzleVQA, each string is one grid row. The symbol \texttt{\#} denotes a wall, \texttt{.} a free cell, and \texttt{G}/\texttt{B} the entrance/exit. The excerpt shows the first and last rows. The intervening rows follow the same encoding.
{\small
\begin{verbatim}
{"grid": ["#######",
          "G...#.#",
          ...,
          "#.....B",
          "#######"]}
\end{verbatim}
}
The program searches adjacent free cells, minimizing visited cells and then turns. It returns 23 cells and eight turns, including the entrance and exit.

\paragraph{Objects and movement constraints in Move Box.}
In this AlgoPuzzleVQA Move Box example, the task is to reach the goal using the fewest box pushes. Locations use zero-based (row, column) coordinates. The wall at (1,3) lies directly above the box at (2,3), preventing the player from standing there to push downward. The omitted wall entries describe the grid boundary.
{\small
\begin{verbatim}
{"grid_size": [6,6],
 "box": [2,3], "goal": [3,4],
 "player": [4,4],
 "walls": [..., [1,3]]}
\end{verbatim}
}
The program tracks both box and player positions and checks player reachability before each push. The minimum is four pushes.

\paragraph{Occupied positions in pattern overlay.}
In VisualPuzzles, each pattern is a set of black-cell positions in a $3\times3$ grid, using zero-based (row, column) coordinates. Here A, B, and C identify the three input patterns, not answer choices.
{\small
\begin{verbatim}
{"patterns": {
   "A": [[0,0],[0,1],[2,1],[2,2]],
   "B": [[0,1],[0,2],[2,0]],
   "C": [[1,1],[1,2],[2,0],[2,1]]}}
\end{verbatim}
}
For aligned overlay, the program takes the union of these sets. A cell covered by multiple patterns is counted once. The union contains eight cells, matching the correct answer.

\paragraph{Discriminative attributes for figure grouping.}
In VisuLogic, \texttt{id} identifies a figure and \texttt{segments} records its number of straight line segments. The six recorded counts are 12, 10, 12, 10, 12, and 10. Two entries are shown below.
{\small
\begin{verbatim}
{"panels": [
   {"id":"1", "segments":12},
   {"id":"2", "segments":10}, ...]}
\end{verbatim}
}
Grouping equal counts yields $\{1,3,5\}$ and $\{2,4,6\}$. The program matches this partition to the correct option B.

\paragraph{Bar heights as visual measurements.}
In PuzzleVQA, each bar has a visible label and top/bottom pixel coordinates. The symbol \texttt{?} marks the missing label. The excerpt retains two bars labeled 2 and the query bar. Height is computed as bottom minus top, so absolute vertical position does not determine the label.
{\small
\begin{verbatim}
{"bars": [ ...,
   {"label":"2", "top":132, "bottom":381},
   ...,
   {"label":"2", "top":128, "bottom":381},
   {"label":"?", "top":132, "bottom":381}]}
\end{verbatim}
}
The program compares the query height with the mean height for each known label. The query is 249 pixels tall. The two bars labeled 2 average 251 pixels, making 2 the nearest match (option A).

\end{document}